\documentclass[10pt,twocolumn,letterpaper]{article}
\usepackage[table,dvipsnames]{xcolor}

\usepackage[pagenumbers]{cvpr} %
\usepackage[disable]{todonotes}

\newcommand{\bff}{\mathbf{f}}\newcommand{\bF}{\mathbf{F}} %

\newcommand{\bI}{\mathbf{I}}

\newcommand{\bp}{\mathbf{p}}\newcommand{\bP}{\mathbf{P}}

\newcommand{\bR}{\mathbf{R}}

\newcommand{\bt}{\mathbf{t}}

\newcommand{\cB}{\mathcal{B}}

\newcommand{\cG}{\mathcal{G}}

\newcommand{\cL}{\mathcal{L}}
\newcommand{\cM}{\mathcal{M}}

\newcommand{\cP}{\mathcal{P}}

\newcommand{\cT}{\mathcal{T}}

\DeclareMathOperator{\argmax}{argmax}

\makeatletter
\DeclareRobustCommand\onedot{\futurelet\@let@token\@onedot}
\def\@onedot{\ifx\@let@token.\else.\null\fi\xspace}
\def\eg{e.g\onedot}

\def\etal{et~al\onedot}

\makeatother

\newcommand{\boldparagraph}[1]{\vspace{0.5em}\noindent{\bf #1.}}

\renewcommand{\paragraph}[1]{\boldparagraph{#1}}

\definecolor{darkgreen}{rgb}{0,0.7,0}
\definecolor{newyellow}{rgb}{1,0.8,0.05}
\definecolor{newgreen}{rgb}{0.2,0.8,0.2}

\def\vis{\text{vis}}
\def\frame{k}
\def\twoD{\text{2D}}
\def\threeD{\text{3D}}
\newcommand{\Fo}[1]{\bF^#1_{o}}
\newcommand{\Fr}[1]{\bF^#1_{r}}

\newcommand{\figteaser}{
\begin{figure}[t]
    \centering
    \includegraphics[width=\linewidth]{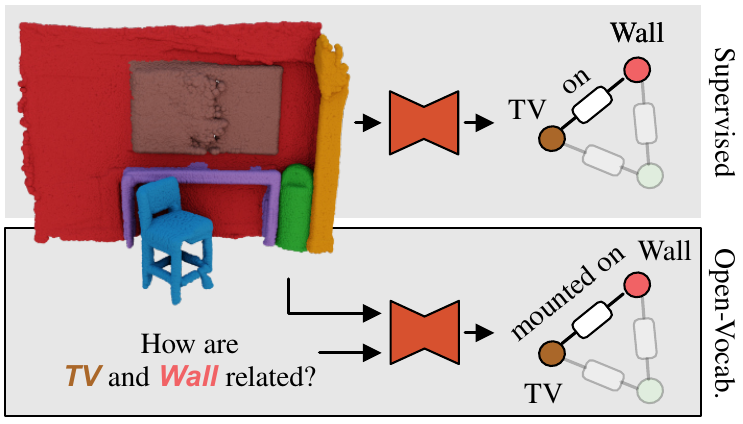}
    \caption{\textbf{\ours.} We present \ours the first approach for learning to predict open-vocabulary 3D scene graphs from 3D point clouds. The advantage of our method is that it can be queried and prompted for any instance in the scene, such as the \textcolor{brown}{\textit{TV}} and \textcolor{red}{\textit{Wall}}, to predict fine-grained semantic descriptions of objects and relationships. By considering all instance pairs in the scene, we can reconstruct a complete explicit open-vocabulary 3D scene graph.}
    \label{fig:teaser}
    \vspace{-1.5em}
\end{figure}}

\newcommand{\figmainmethod}{
\begin{figure*}[t]
    \centering
    \includegraphics[width=\linewidth]{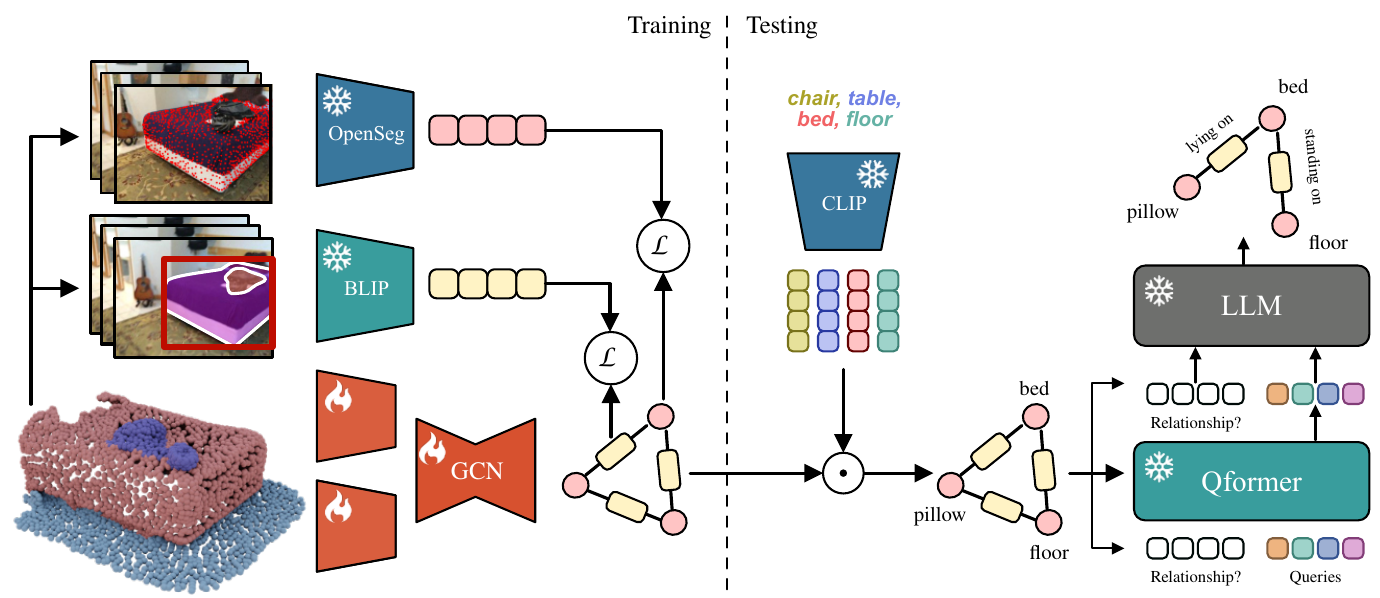}
    \caption{\textbf{\ours overview.} Given a point cloud and RGB-D images with their poses, we distill the knowledge of two vision-language models into our GNN. The nodes are supervised by the embedding of OpenSeg \cite{ghiasi_2022_eccv} and the edges are supervised by the embedding of the InstructBLIP \cite{dai2023instructblip} vision encoder. At inference time, we first compute the cosine similarity between object queries encoded by CLIP \cite{radford_2021_PMLR} and our distilled 3D node features to infer the object classes. Then we use the edge embedding as well as the inferred object classes to predict relationships for pairs of objects using a Qformer \& LLM from InstructBLIP.}
    \vspace{-1em}
    \label{fig:main_method}
\end{figure*}
}

\newcommand{\figframeselect}{
\begin{figure*}[t]
    \centering
    \includegraphics[width=\linewidth]{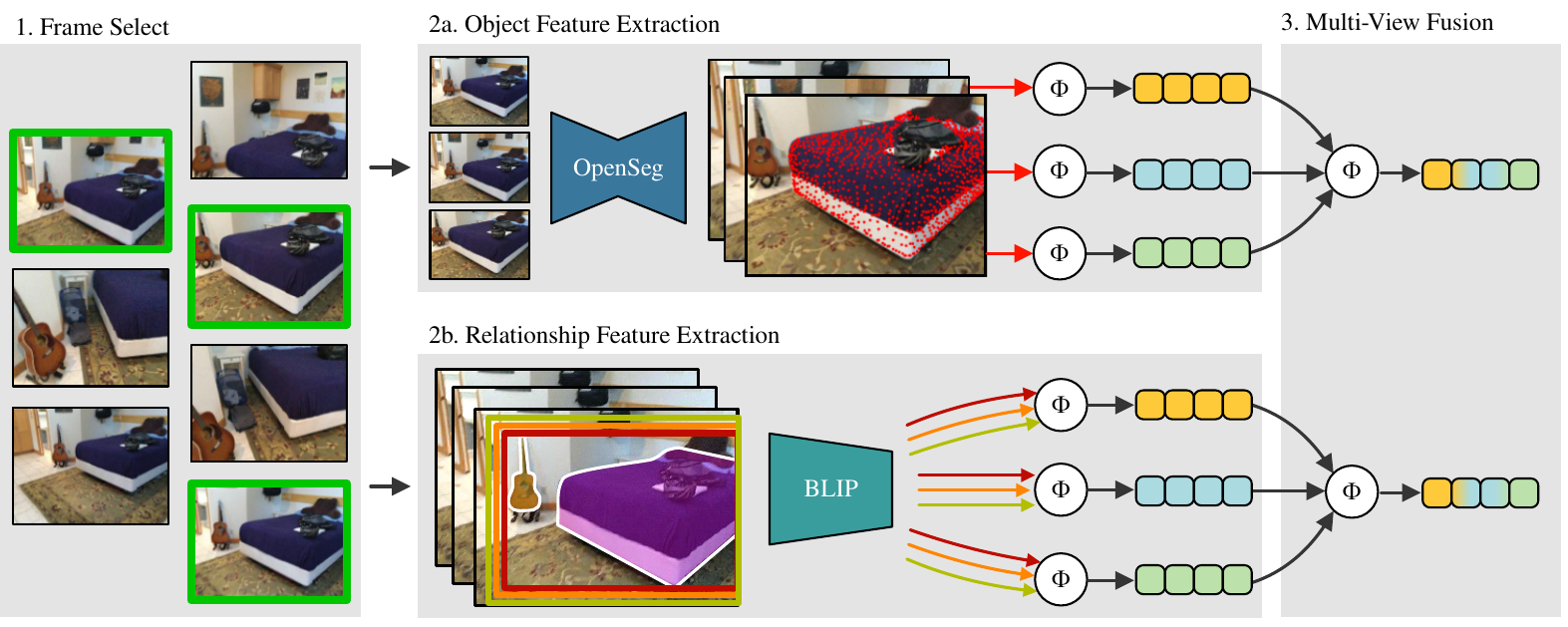}
    \caption{\textbf{Supervision feature extraction.} For each instance in the 3D point cloud, we select the top k frames for object and predicate supervision. For objects, we encode the frames using OpenSeg \cite{ghiasi_2022_eccv} and aggregate the computed features over the projected points. For predicates, we identify object pairs in the frame, crop the image at multiple scales and compute the image feature with the BLIP \cite{dai2023instructblip} image encoder. The features are aggregated over all crops. Finally, both object and predicate features are fused across the multiple views.}
    \vspace{-1em}
    \label{fig:frame_select}
\end{figure*}
}

\newcommand{\figtripletloc}{
\begin{figure}[t]
    \centering
    \includegraphics[width=\linewidth]{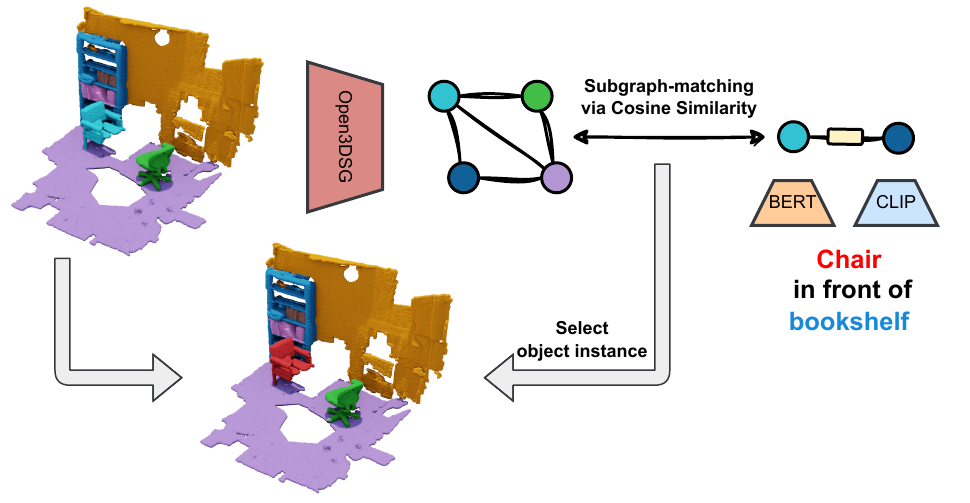}
    \caption{\textbf{Application: Object localization via triplet description.} Using our open-vocabulary approach, we can localize object instances in the 3D point cloud given a relationship description of the object instance.}
    \label{fig:triplet_loc}
\end{figure}
}

\newcommand{\figweightreasoning}{
\begin{figure}[t]
    \centering
    \includegraphics[width=\linewidth]{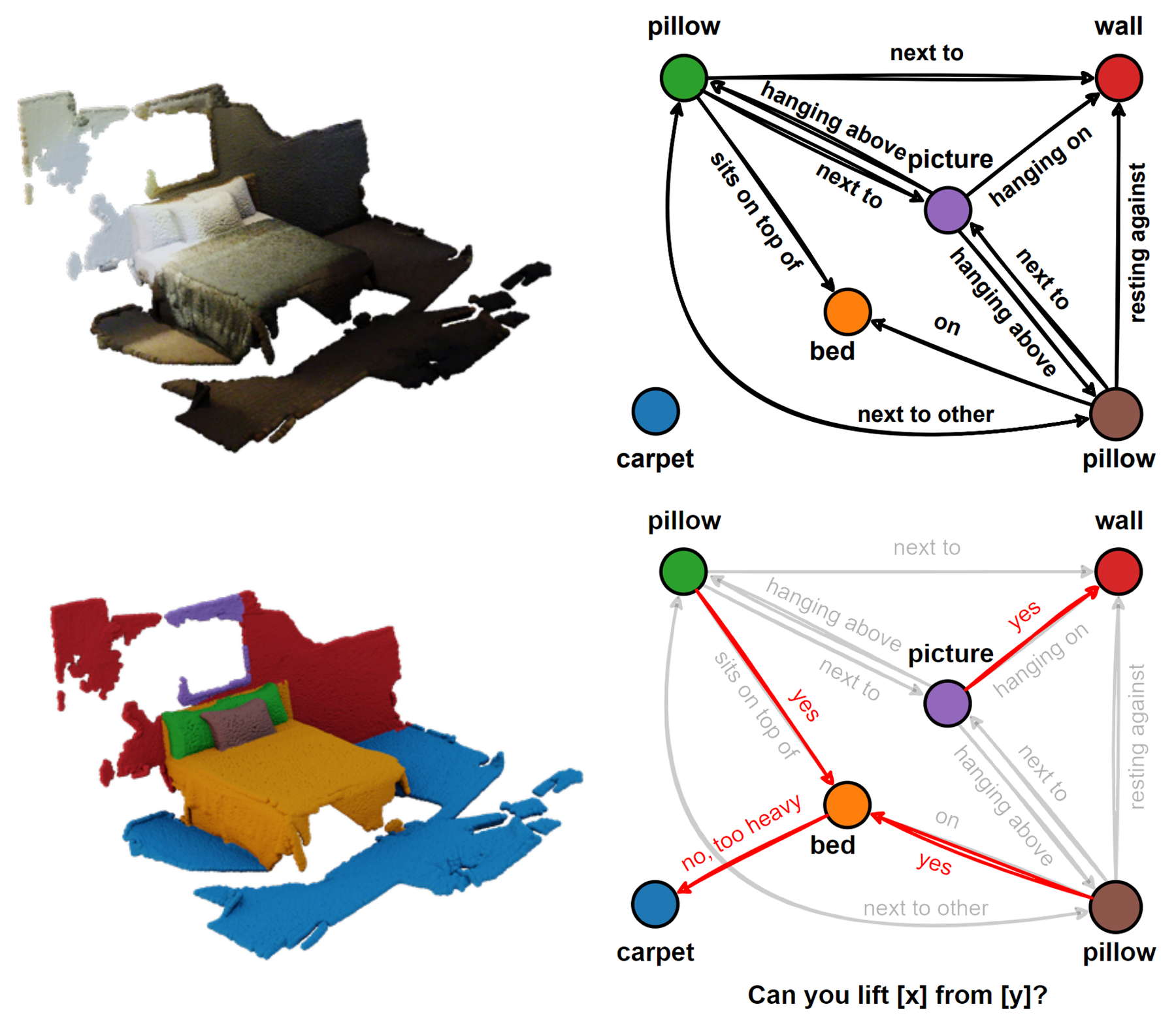}
    \caption{\textbf{Application: Reasoning over object affordances.} Using our open-vocabulary approach, we reason about the affordances of objects by for instance prompting the LLM to output whether an object can be lifted from the other.}
    \label{fig:weight_reasoning}
\end{figure}
}

\newcommand{\figmaterial}{
\begin{figure}[t]
    \centering
    \includegraphics[width=\linewidth]{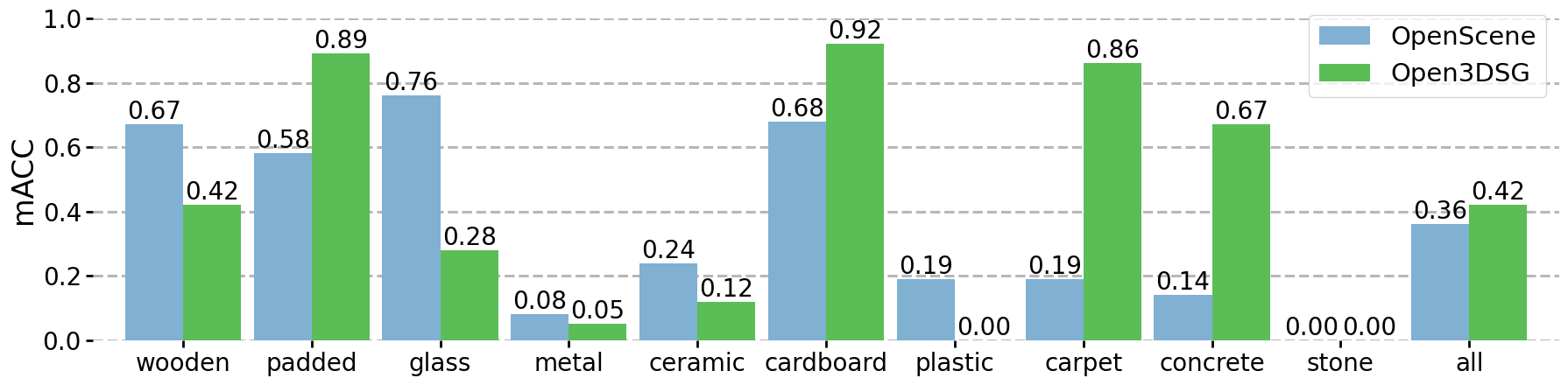}
    \caption{\textbf{Application: Material prediction.} Using our open-vocabulary approach, we predict the material of objects without explicit training. We compare against OpenScene \cite{Peng2023OpenScene}.}
    \label{fig:material}
\end{figure}
}

\newcommand{\figquality}{
\begin{figure*}[t]
    \centering
    \begin{subfigure}[c]{0.24\textwidth}
         \centering
         \includegraphics[width=\textwidth]{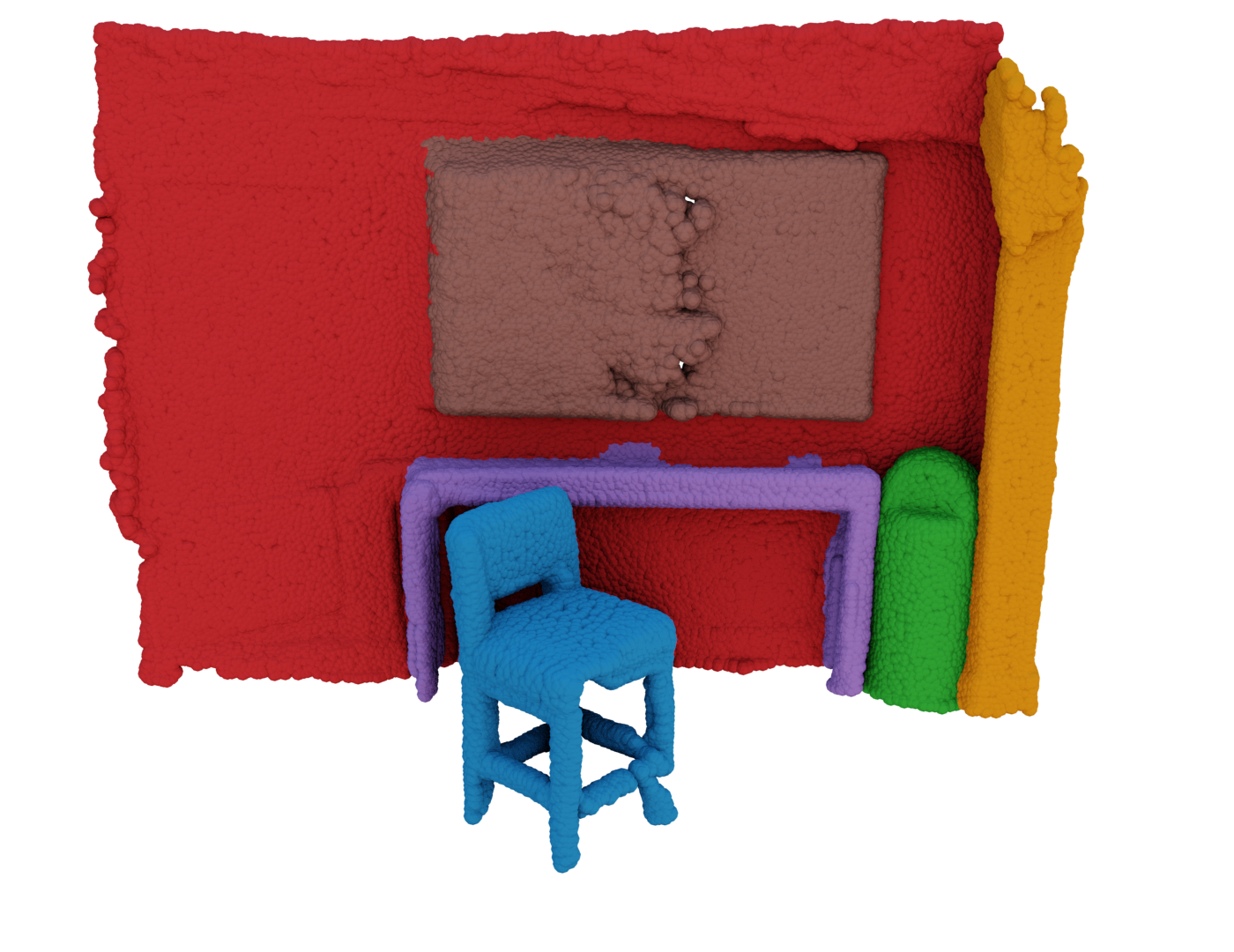}
     \end{subfigure}
     \begin{subfigure}[c]{0.25\textwidth}
         \centering
         \includegraphics[width=\textwidth]{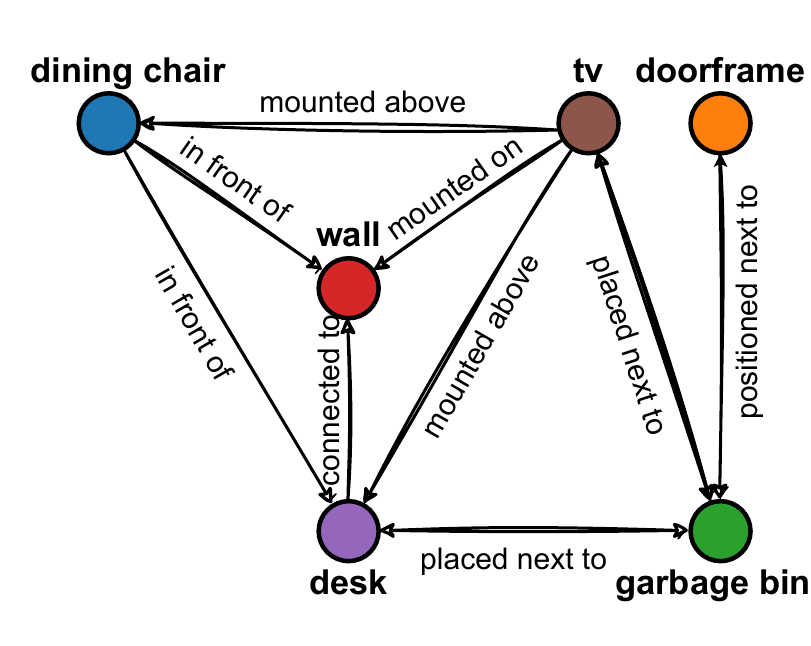}
     \end{subfigure}
     \begin{subfigure}[c]{0.24\textwidth}
         \centering
         \includegraphics[width=\textwidth]{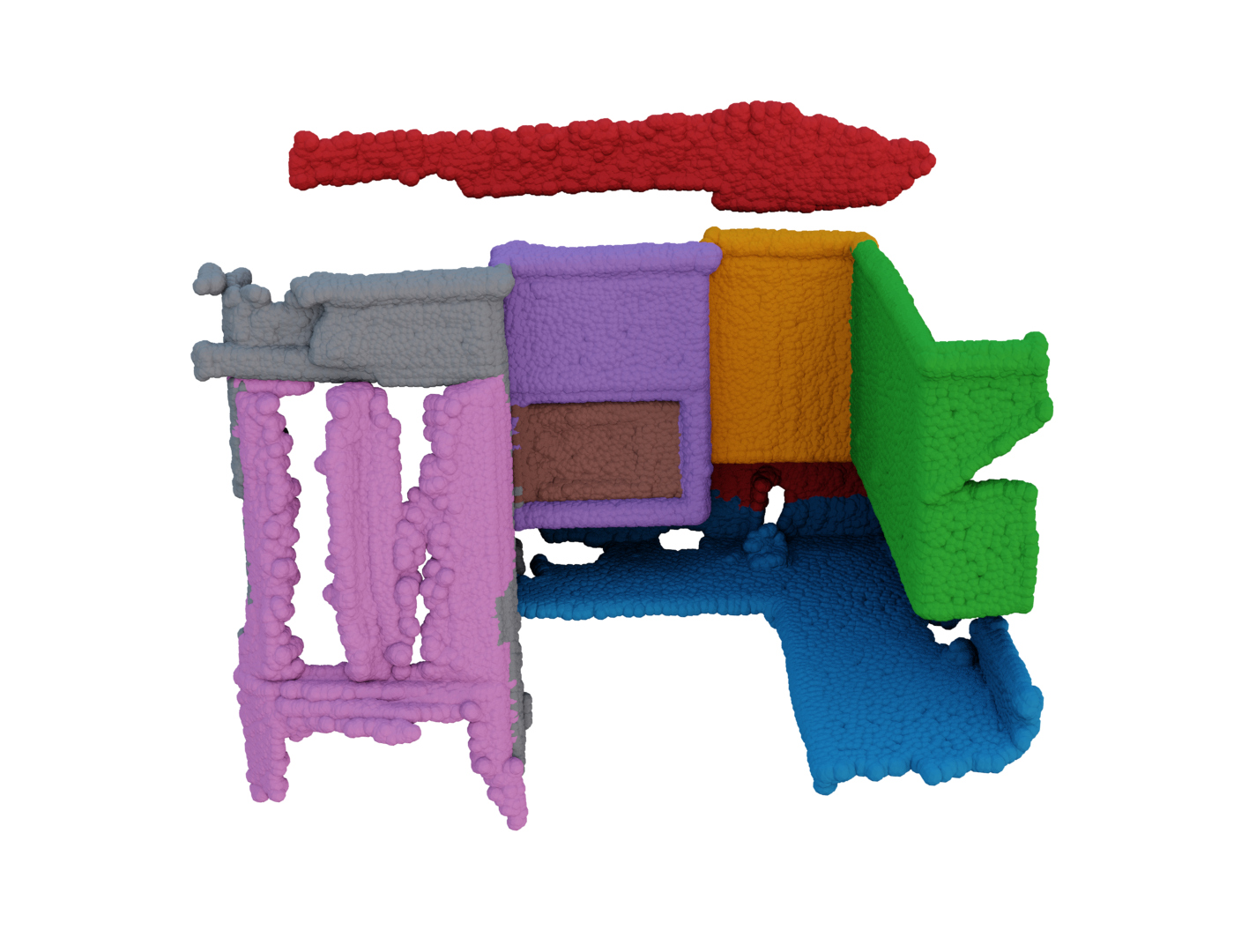}
     \end{subfigure}
     \begin{subfigure}[c]{0.25\textwidth}
         \centering
         \includegraphics[width=\textwidth]{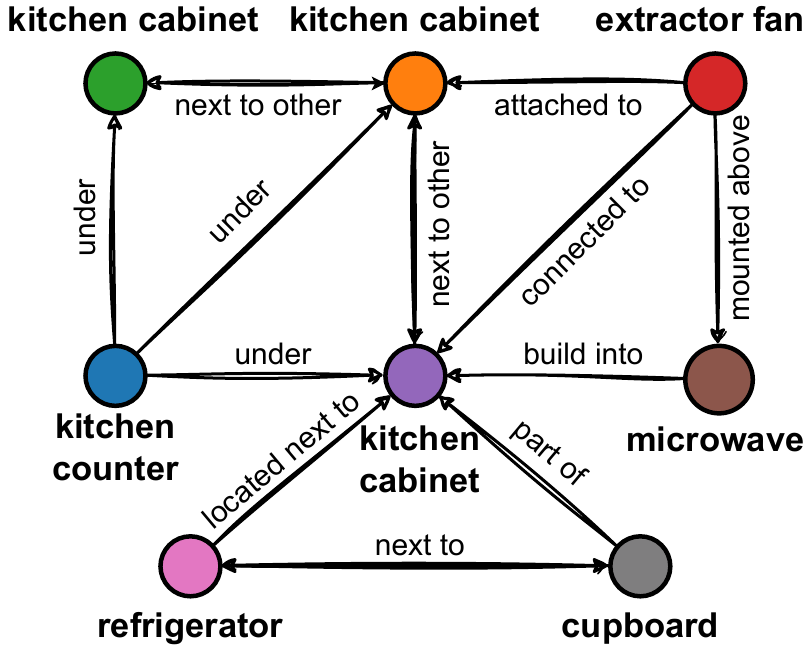}
     \end{subfigure}

    \caption{\textbf{Qualitative open-vocabulary 3D scene graph predictions.} We show the top-1 predictions on ScanNet \cite{Dai_2017_CVPR} from \ours. The nodes are queried using the 3DSSG \cite{Wald_2020_CVPR} 160 class label set, while the edges are generated directly from the graph-conditioned LLM.}
    \label{fig:qualitative_results}
    \vspace{-1em}
\end{figure*}
}

\newcommand{\tabmaineval}{
\begin{table}[t]
\tabcolsep=0.2mm
\centering
\small
\begin{tabular}{@{}lcccccc@{}}
\toprule
       & \multicolumn{2}{c}{Object} & \multicolumn{2}{c}{Predicate} & \multicolumn{2}{c}{Relationship} \\
       \cmidrule(lr){2-3}\cmidrule(lr){4-5}\cmidrule(lr){6-7}
      Method & R@5                  & R@10                 & R@3                 & R@5                & R@50                 & R@100                \\
      \midrule
\multicolumn{3}{l}{Fully-supervised} \\
\rowcolor{lightgrey}
3DSSG \cite{Wald_2020_CVPR} & 0.68                 & 0.78                 & 0.89                & 0.93               & 0.40                 & 0.66                 \\ 
\rowcolor{lightgrey}
SGFN \cite{Wu_2021_CVPR} & 0.70 & 0.80 & 0.97 & \textbf{0.99} & 0.85 & 0.87 \\
\rowcolor{lightgrey}
SGRec3D \cite{koch2023sgrec3d} &\textbf{0.80} & \textbf{0.87} & 0.97 & \textbf{0.99} & 0.89 & 0.91 \\
\rowcolor{lightgrey}
VL-SAT \cite{wang2023vl} &0.78 & 0.86 & \textbf{0.98} & \textbf{0.99} & \textbf{0.90} & \textbf{0.93} \\
\midrule
\multicolumn{3}{l}{\textit{Zero-shot open-vocabulary}}  \\
CLIP (naive) \cite{radford_2021_PMLR} & 0.35 & 0.42  & 0.09 & 0.19 & 0.02 & 0.04 \\
OpenSeg \cite{ghiasi_2022_eccv} + CLIP \cite{radford_2021_PMLR} & 0.38 & 0.45  & 0.10 & 0.23 & 0.05 & 0.07 \\
OpenSeg \cite{ghiasi_2022_eccv} + NCLIP \cite{yuksekgonul2023when} & 0.38 & 0.45  & 0.10 & 0.20 & 0.05 & 0.08 \\
OpenSeg \cite{ghiasi_2022_eccv} + Cap. \cite{li2023blip} & 0.38 & 0.45 & 0.50 & 0.58 & 0.30 & 0.32 \\

\textbf{\ours (Ours}) & \textbf{0.57} & \textbf{0.68}  & \textbf{0.63} & \textbf{0.70} & \textbf{0.64} & \textbf{0.66} \\
\bottomrule
\end{tabular}
\caption{\textbf{Closed-vocabulary evaluation on 3DSSG.} We compare our method with both zero-shot and fully-supervised baselines for 3D scene graph prediction. Overall, the zero-shot approaches perform worse than the fully-supervised methods. However, \ours achieves comparable results to the first supervised 3D scene graph prediction method 3DSSG.}
\label{tab:main_evaluation}
\vspace{-1em}
\end{table}
}

\newcommand{\tabablation}{
\begin{table}[t]
\tabcolsep=2.1mm
\centering
\footnotesize
\begin{tabular}{@{}lcccc@{}}
\toprule
       & \multicolumn{2}{c}{Object} & \multicolumn{2}{c}{Predicate}\\
       \cmidrule(lr){2-3}\cmidrule(lr){4-5}
       & R@5            & mR@5        & R@3                      & mR@3                    \\
      \midrule

      \ours 2D & 0.37 & 0.37 & 0.67 & 0.19\\
      \ours 3D & 0.46 & 0.25 & 0.60 & 0.33\\
      \ours 2D-3D & 0.57 & 0.45 & 0.63 & 0.37\\
      \midrule

      \ours 2D-3D w/ CLIP & 0.48 & 0.32 & 0.59 & 0.32 \\
      \ours 2D-3D + GT Objs & 1.00 & 1.00 & 0.64 & 0.38 \\
      \ours 2D-3D + Supv. Rels. & 0.59 & 0.46 & 0.76 & 0.44 \\
\bottomrule
\end{tabular}
\caption{\textbf{Ablation study.} 3D scene graph prediction with different input modalities, object VLM, privileged ground-truth information and supervised fine-tuning.}
\label{tab:mrec}
\vspace{-0.5cm}
\end{table}
}

\newcommand{\tabheadbodytailbaselines}{
\begin{table}[t]
\tabcolsep=0.7mm
\centering
\small
\begin{tabular}{@{}lcccccc@{}}
\toprule
      & & Labels & Head & Body & Tail & All\\
       \midrule
       \multirow{4}{*}{Objects R@5} & 
       3DSSG \cite{Wald_2020_CVPR} & \footnotesize{$10^5$} & 0.88 & 0.45 & 0.06 & 0.30 \\
       & SGRec3D \cite{koch2023sgrec3d} & \footnotesize{$10^5$} & \textbf{0.92} & \textbf{0.78} & 0.24 & 0.45 \\
       & VL-SAT \cite{wang2023vl}& \footnotesize{$10^5$} & \textbf{0.92} & 0.73 & 0.31 & \textbf{0.46} \\
       \cmidrule{4-7}
       & \ours & 0 & 0.60 & 0.50 & \textbf{0.42} & 0.45 \\
       \midrule
       \midrule
       \multirow{4}{*}{Predicates R@3} & 
       3DSSG \cite{Wald_2020_CVPR}  & \footnotesize{$10^5$} & 0.94 & 0.83 & 0.41 & 0.57 \\
       & SGRec3D \cite{koch2023sgrec3d}  & \footnotesize{$10^5$} & 0.97 & \textbf{0.96} & \textbf{0.65} & 0.69 \\
       & VL-SAT \cite{wang2023vl}  & \footnotesize{$10^5$} & \textbf{0.99} & 0.94 & 0.58 & \textbf{0.75} \\
       \cmidrule{4-7}
       & \ours & 0 & 0.38 & 0.29 & 0.57 & 0.37 \\ %
\bottomrule
\end{tabular}
\caption{\textbf{Frequency based class evaluation.} Here we compare the prediction performances for objects and predicates based on their frequency in the training set. Even though the fully-supervised
approaches are trained specifically on this dataset, we can handle the less-common / long-tail classes much better.}
\label{tab:headbodytail}
\vspace{-1em}
\end{table}
}

\newcommand{\tabclasscompare}{
\begin{table}[h]
\tabcolsep=1.5mm
\centering
\small
\begin{tabular}{@{}lccc@{}}
\toprule
      & 3DSSG & SGRec3D & \ours  \\
       \midrule
       \multicolumn{3}{l}{\textit{Objects R@5}} \\
       cabinet / kitchen cabinet  & 0.39 / 0.33  &  0.67 / 0.87  &  0.39 / 0.94  \\
       chair / dining chair &  0.98 / 0.00  & 0.94 / 0.00  &  0.48 / 1.00  \\
       table / bedside table  &  0.60 / 0.00 &  0.90 / 0.25  &  0.37 / 1.00  \\
       \midrule
       \multicolumn{3}{l}{\textit{Predicates R@3}} \\
       standing on & 0.73 & 0.95 & 0.86 \\
       covering & 0.00 & 0.00 & 0.24\\
       belonging to & 0.48 & 0.65 & 0.91\\
\bottomrule
\end{tabular}
\caption{\textbf{Semantic awareness.} While fully-supervised methods such as 3DSSG [44] and SGRec3D [22] produce overall good results, their performance on difficult, rare, and semantically descriptive classes remains low. In contrast our open-vocabulary approach excels at semantically descriptive classes.}
\label{tab:class_study}
\end{table}
}

\newcommand{\tabvlcheck}{
\begin{table}[h]
\tabcolsep=6.6mm
\centering
\small
\begin{tabular}{@{}lccc@{}}
\toprule
      & top-1 & top-3 & top-5  \\
       \midrule
       Random chance & 0.04  & 0.12 & 0.19  \\
       CLIP (ViT-L/14) & 0.12 & 0.30 & 0.42 \\
       NegCLIP & 0.14 & 0.35 & 0.48 \\
       SigLIP & 0.11 & 0.27 & 0.37 \\
\bottomrule
\end{tabular}
\caption{\textbf{VL-Checklist Relation.} We evaluate the embedded relationship knowledge of the current state of contrastively pre-trained VLMs on an adapted benchmark from [52]. Results are reported for whether the VLM scores the correct predicate in the top-1, \mbox{top-3}, or top-5.}
\label{tab:vgr}
\end{table}
}

\usepackage{xcolor}
\definecolor{turquoise}{cmyk}{0.65,0,0.1,0.3}
\definecolor{purple}{rgb}{0.65,0,0.65}
\definecolor{dark_green}{rgb}{0, 0.5, 0}
\definecolor{orange}{rgb}{0.8, 0.6, 0.2}
\definecolor{red}{rgb}{0.8, 0.2, 0.2}
\definecolor{darkred}{rgb}{0.6, 0.1, 0.05}
\definecolor{blueish}{rgb}{0.0, 0.3, .6}
\definecolor{lightgrey}{rgb}{0.9, 0.9, .9}
\definecolor{lightred}{rgb}{0.9, 0.7, .7}

\definecolor{pink}{rgb}{1, 0, 1}
\definecolor{greyblue}{rgb}{0.25, 0.25, 1}

\definecolor{thirdbestcolor}{rgb}{1,1, 0.6}
\definecolor{secondbestcolor}{rgb}{1, 0.9, 0.6}
\definecolor{firstbestcolor}{rgb}{1, 0.6, 0.6}

\usepackage{lipsum}  
\usepackage{nicefrac}
\usepackage{multirow} 
\usepackage{pifont}

\usepackage[
final,tracking=true,kerning=true,spacing=true,factor=1100,stretch=10,shrink=10]{microtype}
\microtypesetup{protrusion=false}
\usepackage[accsupp]{axessibility}
\pdfoutput=1

\definecolor{cvprblue}{rgb}{0.21,0.49,0.74}
\usepackage[pagebackref,breaklinks,colorlinks,citecolor=cvprblue]{hyperref}

\def\paperID{12206} %
\def\confName{CVPR}
\def\confYear{2024}

\newcommand{\ours}{Open3DSG\xspace}
\newcommand{\mytitle}{\ours: Open-Vocabulary 3D Scene Graphs from Point Clouds\\ with Queryable Objects and Open-Set Relationships}

\title{\mytitle}

\author{
Sebastian Koch$^{1,2,3}$\qquad Narunas Vaskevicius$^{1,2}$\qquad Mirco Colosi$^2$\qquad \\Pedro Hermosilla$^4$\qquad Timo Ropinski$^3$ \vspace{0.05cm}\\ 
{\small $^1$\text{Bosch Center for Artificial Intelligence}\quad $^2$Robert Bosch Corporate Research} \quad {\small $^3$\text{University of Ulm} \quad $^4$ TU Vienna}
\\{\small\href{https://kochsebastian.com/open3dsg}{kochsebastian.com/open3dsg}}
}
\begin{document}
\maketitle
\begin{abstract}
Current approaches for 3D scene graph prediction rely on labeled datasets to train models for a fixed set of known object classes and relationship categories. 
We present \ours, an alternative approach to learn 3D scene graph prediction in an open world without requiring labeled scene graph data. 
We co-embed the features from a 3D scene graph prediction backbone with the feature space of powerful open world 2D vision language foundation models. 
This enables us to predict 3D scene graphs from 3D point clouds in a zero-shot manner by querying object classes from an open vocabulary and predicting the inter-object relationships from a grounded LLM with scene graph features and queried object classes as context. 
\ours is the first 3D point cloud method to predict not only explicit open-vocabulary object classes, but also open-set relationships that are not limited to a predefined label set, making it possible to express rare as well as specific objects and relationships in the predicted 3D scene graph. 
Our experiments show that \ours is effective at predicting arbitrary object classes as well as their complex inter-object relationships describing spatial, supportive, semantic and comparative relationships.
\end{abstract}
    
\vspace{-1em}
\section{Introduction}
\label{sec:intro}
\figteaser

3D scene graphs are an emergent graph-based representation facilitating various 3D scene understanding tasks. 
In contrast to other more object-centric 3D scene representations, the key advantage of 3D scene graphs is the ability to also represent relationships between scene entities, such as for instance objects in a room. 
These relationships can be useful for a variety of different downstream tasks in computer vision or robotics, such as place recognition, change detection, task planning and more \cite{Wald_2020_CVPR, looper22vsg,rana2023sayplan,Agia_2022_PMLR,zhai2023sg}. 
However, the exploitation of 3D scene graphs is limited by their availability.%

Given their complexity and high-level abstraction, 3D scene graphs are hard to predict by learned models. The state-of-the-art (SOTA) methods for 3D scene graph prediction are limited to a fixed set of object and relationship labels provided by small-scale datasets. This reduces their effectiveness in downstream applications, which often require semantic reasoning on concepts extending beyond a rather narrow scope of training data. Furthermore, one of the most useful properties of scene graphs is their ability to represent relationships between scene entities. There are multiple ways of describing a relationship between two objects, e.g. spatial, comparative, semantic, etc. The relevance of the type of relationship is dictated by the downstream task. However, in a closed-set supervised training setting this choice is made and fixed in advance.

Open-vocabulary 3D scene understanding methods propose a solution towards these challenges by training a model not on a fixed label set but rather aligning the 3D model with 2D foundation models~\cite{Peng2023OpenScene,takmaz2023openmask3d,kerr2023lerf,hegde2023clip,conceptfusion,ha2022semantic}. 
By doing so, \eg with foundation models such as CLIP \cite{radford_2021_PMLR},
the 3D model can express nearly the same broad vocabulary that these vision language models (VLMs) were trained on.
However, while these 2D models are very capable of predicting single objects or higher-level concepts, they do not perform well in modeling compositional structures such as relationships or attributes \cite{yuksekgonul2023when,yamada2022lemons}.
This limitation makes it challenging to adopt 2D VLMs for scene graph predictions where compositional relationships are the core part. 

In this paper, we demonstrate that intuitive CLIP-like approaches are ill-suited for open-vocabulary relationship prediction.
To this end, our key idea is to combine the advantages of VLMs with large language models (LLMs), that have proven to be better at understanding compositional concepts \cite{hsieh2023sugarcrepe}, to predict open-vocabulary 3D scene graphs.

\noindent We highlight the following three contributions:

\begin{itemize}
    \item We are the first to present a method to create an interactive graph representation of a scene from a 3D point cloud, which can be queried for objects and prompted for relationships during inference time.
    \item We show how such a representation can be converted into an explicit open-vocabulary 3D scene graph. Thus effectively proposing the first open-vocabulary scene graph prediction approach from 3D point cloud data.
    \item Our proposed approach shows promising results on the closed-set benchmark 3DSSG \cite{Wald_2020_CVPR}, proving success in modeling compositional concepts in an open-vocabulary manner.\looseness=-1
\end{itemize}

\section{Related Work}
\vspace{-0.5em}
\boldparagraph{3D scene graph prediction} 
3D scene graphs were first proposed by Armeni \etal \cite{Armeni_2019_ICCV} as a hierarchical structure to combine entities such as buildings, rooms, objects and cameras into a unified structure. Following their inception, subsequent works improved upon the estimation of such hierarchical 3D scene graphs for large-scale environments \cite{Rosinol_2020_RSS,hughes2022hydra,Rosinol_2021_SAGE}. Other 3D scene graph approaches rather focus on predicting local semantic inter-object relationships and building a graph of objects~\cite{Wald_2020_CVPR,Zhang_2021_CVPR,wald2022learning,Zhang_2021_Neurips,Wu_2021_CVPR,Wu_2023_CVPR,wang2023vl,koch2023lang3dsg}. The applications of these 3D scene graphs are plentiful, with uses in aligning 3D scans \cite{sarkar2023sgaligner}, reconstructing and generating 3D scenes \cite{Dhamo_2021_ICCV,koch2023sgrec3d}, forecasting scene change \cite{looper22vsg}, or even task planning over 3D scene graphs \cite{Agia_2022_PMLR,rana2023sayplan}.
However, none of these approaches consider the topic of open vocabulary in the context of 3D scene graphs. Cheng \etal are the first to model an implicit scene graph representation for planning in navigation tasks which they call OVSG \cite{chang2023context} -- an open-vocabulary 3D scene graph model -- however they do not predict any open-vocabulary relationships from sensor data and are reliant on human descriptions which are encoded in the scene graph using a language model for open-vocabulary lookup and matching. Another approach to explore open-vocabulary 3D scene graphs is ConceptGraphs~\cite{conceptgraphs} which is concurrent work to ours. ConceptGraphs utilizes 2D VLMs and captioning models to predict scene graphs with queryable nodes and stored summarized image captions for edges. However, they do not provide extensive evaluations for their predicted scene graphs, limiting themselves to a qualitative evaluation of spatial relationships with Amazon Turk.
We identify that the core difference of our approach to ConceptGraphs and OVSG, is that we learn to predict 3D scene graphs directly from raw point clouds, which brings numerous advantages such as being able to predict 3D scene graphs at test time without requiring inference from computationally expensive VLMs and when only 3D scans are available. We also predict explicit semantic relationships as part of our method and do not have to store multiple captions per edge that describe the relationship.

\vspace{-0.60em}
\boldparagraph{Open-vocabulary 3D scene understanding}
The recent success of 2D vision language models as open-vocabulary methods such as CLIP \cite{radford_2021_PMLR}, ALIGN \cite{pmlr-v139-jia21b}, or ImageBind~\cite{Girdhar_2023_CVPR} have motivated the process of adapting these foundation models for 3D scene understanding tasks such as semantic/instance segmentation or 3D open-vocabulary detection. One of the earliest lines of approaches \cite{Zhang_2022_CVPR,Zhu_2023_ICCV,hegde2023clip,ha2022semantic} and also ConceptGraphs \cite{conceptgraphs} explore annotation-free 3D recognition by combining CLIP with a 3D detection head using available RGB-D images with known poses. However, these approaches can suffer from inaccurate 2D-3D projections and occlusion artifacts.
Furthermore, RGB-D images with known poses are not always available. 
Therefore, more recently approaches such as OpenScene \cite{Peng2023OpenScene}, LERF \cite{kerr2023lerf} and others \cite{conceptfusion,zhang2023clip,takmaz2023openmask3d,lu2023ovir} aim to distill the knowledge of those 2D vision language models into a 3D architecture with the advantage that these approaches do not rely on available 2D images when performing inference on 3D data. After the distillation, these approaches demonstrate impressive open-vocabulary results and unique abilities such as localizing rare objects in large 3D scans. However, their accuracy on closed vocabulary benchmarks still falls short of fully-supervised methods that are specifically trained on one dataset.

However, in contrast to our goal, none of these 3D scene understanding approaches has attempted to model 3D relationships which are hard to learn and distill based on their compositional nature. 

\figmainmethod
\vspace{-0.60em}
\boldparagraph{Compositionality in vision-language models}
\label{sec:related_compositional}
While vision-language models show impressive performances in zero-shot image retrieval or image classification \cite{radford_2021_PMLR,pmlr-v139-jia21b, zhai2023sigmoid, tschannen2023image, Girdhar_2023_CVPR}%
, they lack complex compositional understanding. Yuksekgonu \etal \cite{yuksekgonul2023when} and Yamada \etal  \cite{yamada2022lemons} identified that contrastive vision-language pre-trained models such as CLIP \cite{radford_2021_PMLR} tend to collapse to a \textit{bag-of-words} representation, which cannot disentangle multi-object concepts. 
To this end, a number of benchmarks have surfaced to examine the compositional reasoning capabilities of current vision language models \cite{yuksekgonul2023when,Ma_2023_CVPR,hsieh2023sugarcrepe,Thrush_2022_CVPR}. Yet, attempts to improve compositional understanding of contrastive vision-language pre-trained models by utilizing additional data, prompting, models, losses and/or hard negatives \cite{yuksekgonul2023when,Cascante-Bonilla_2023_ICCV,nayak2022learning,ray2023cola,Doveh_2023_CVPR,singh2023coarse} yield only marginal improvements on these benchmarks. Furthermore, it is unclear whether these models achieve these improvements by actually acquiring compositional understanding or by exploiting biases in these benchmarks as indicated in \cite{hsieh2023sugarcrepe}.

Predicting relationships in a scene graph requires compositional understanding. In this paper, we approach this problem by shifting from a discriminative zero-shot approach to a generative approach using an LLM.
\todo[inline]{explain better how we are dealing with the problem and why we are doing it like this}

\figframeselect
\section{Method}
The overall goal of our approach is to distill the knowledge of 2D vision-language models into a 3D graph neural network (GNN) to predict open-vocabulary 3D scene graphs in a 2-step process. 
We first construct an initial graph representation (\cref{sec:graph_construction}), and in parallel, we extract vision-language features from aligned 2D images (\cref{sec:frame_select}).
These features are then aligned to the ones extracted via the 3D GNN (\cref{sec:distillation}), so that we can predict the same language-aligned features from 3D data only. 
At inference time, we perform a two-step prediction for objects and relationships. First, we predict object classes via a cosine similarity between the distilled features and open-vocabulary queries encoded by CLIP~\cite{radford_2021_PMLR}. Then, we predict inter-object relationships by providing the learned relationship feature vector and the predicted object classes as context for a LLM (\cref{sec:prediction}).
An overview of our method is shown in \cref{fig:main_method}.

\todo[inline]{mention that we provide model infos in the supplementary, such as how many layers, which OpenSeg/CLIP model, which InstructBLIP model, embedding size, learning rate, trained epochs, hardware}
\subsection{Scene graph construction}
\label{sec:graph_construction}

Given a point cloud $\cP$ of a scene with class-agnostic instance segmentation $\cM$ provided by an off-the-shelf instance segmentation method such as Mask3D~\cite{Schult_2022_CORR} or the dataset itself, we extract each object point cloud $\cP_i$ containing instance $i$ using the mask $\cM_i$. 
Further, we extract point clouds $\cP_{ij}$ of the instance pair \mbox{$\langle i, j \rangle \in |\cM|\times |\cM|$}, by selecting all points falling within the union of their respective bounding boxes $\cB_{ij} = \cB_{i} \cup \cB_{j}$.

We construct an initial graph with node features $\phi_n$ and edge features $\phi_e$. Each point set $\cP_i$ is fed into a shared PointNet~\cite{Qi_2017_CVPR} to extract features for object nodes. 
Every point set $\cP_{ij}$ is concatenated with a mask which is equal to $1$ if the point corresponds to object $i$, $2$ if the object corresponds to object $j$, and $0$ otherwise. 
The concatenated feature vector is then fed into another shared PointNet to extract features for predicate edges. 

The extracted node and edge features are then arranged as triplets \mbox{$t_{ij} = \langle \phi_{n,i},\phi_{e,ij},\phi_{n,j} \rangle$} in a graph structure. 
This initial feature graph is passed into a GNN that processes the triplets $t_{ij}$ and propagates the information through the graph \looseness=-1
\begin{equation}
    \small \phi_{n,i}^{(k)}, \phi_{e,ij}^{(k)}, \phi_{n,j}^{(k)} = \text{G}(\phi_{n,i},\phi_{e,ij},\phi_{n,j})
\end{equation}
where $\text{G}(\cdot)$ is a GNN and $\phi_{n,i}^{(k)}, \phi_{e,ij}^{(k)}, \phi_{n,j}^{(k)}$ are the refined features after $k$ iterations of the GNN.

\subsection{2D feature extraction}
\label{sec:frame_select}
\vspace*{-0.3cm}
\boldparagraph{Frame selection}
The first step for aligning our 3D GNN with the 2D vision-language models is to extract 2D features from the available 2D images and project them onto the constructed 3D feature graph. 
Selecting high-quality frames where the desired objects are visible is crucial to obtain robust and high-quality features. 
To achieve this, we utilize the same class-agnostic instance mask already used before to select a small subset of frames 
containing each pair of objects to be aligned with nodes and edges in the produced graph.\looseness=-1

To estimate whether an object of instance $i$ is visible in frame $\frame$, we use the camera's intrinsic $\bI$ and extrinsics $(\bR_\frame|\bt_\frame)$ to project all points $\cP_i$ onto the image plane of frame $\frame$. We define the projection of a single point $p_i$ belonging to instance $i$ projected into frame $\frame$ with $ \bp_{i\frame} = (u,v,w)^T=\text{proj}_\frame(p_i)=\bI\cdot(\bR_\frame|\bt_\frame) \cdot p_i$ where we represent $p_i$ in homogeneous coordinates. 
We consider a point falling into the image plane if $\nicefrac{u}{w}$ falls in the interval $[0,W-1]$ and likewise if $\nicefrac{v}{w}$ falls in the interval $[0,H-1]$, where W and H are the image width and height dimension respectively.
Furthermore, 
we discard each point $\bp_{ik}$ that is occluded from the point of view of frame $\frame$, for which the inequality $w-d_\frame>t_{\text{occ}}$ is satisfied, where $d_\frame$ is the measured depth for pixel $(u,v)$, $w$ is the estimated depth of the object instance for the same pixel, and $t_{\text{occ}}$ is a fixed threshold hyperparameter.
We denote the set of projected points passing the validity checks as $\bP_{i\frame}$.
Subsequently, we compute a visibility percentage as\\
\begin{equation}
    \vis(i,\frame) = \frac{|\bP_{i\frame}|}{|\cP_i|}
    \label{eq:visibility_fac}
\end{equation}
expressing the ratio of object points that are successfully projected onto the image frame. From the projected points, we can estimate their bounding box in the image as \mbox{$box_{i\frame} = [\min_x(\bP_{i\frame}),\min_y(\bP_{i\frame}),\max_x(\bP_{i\frame}),\max_y(\bP_{i\frame})]$}.
Following this projection routine, each object instance $i$ can be projected onto multiple frames. To ensure high-quality visual features, we choose a subset of high-quality frames by rejecting low-quality ones based on the condition
\begin{equation}
    \vis(i,\frame) > t_{\vis} \lor \text{A}(box_{i\frame}) > t_{box}
    \label{eq:cirteria}
\end{equation}
where $t_{\vis}$ and $t_{box}$ are hyperparameters, and $\text{A}(\cdot)$ computes the area of the given bounding box. 
We consider the bounding box area as an additional condition since large objects, such as \textit{floor} or \textit{wall}, 
might cover a huge portion of the scene, leading to a low visibility percentage for the current frame. 
In the end, we choose the top-k frames with the highest quality.
For relationship frame selection, the process is similar, but we consider two object instances $\cP_i$ and $\cP_j$ simultaneously and a candidate frame has to satisfy \cref{eq:cirteria} for both objects.
The process of selecting both object and relationship frames is shown in \cref{fig:frame_select} box $1$. 
\todo[inline]{make sure the figure box reference is up to data with new figure}

\boldparagraph{Object feature computation}
In order to achieve a coherent language-aligned object feature, we decide to leverage a VLM and collect the extracted features in a single representation. 
We choose OpenSeg \cite{ghiasi_2022_eccv} over CLIP \cite{radford_2021_PMLR}, since the latter returns a global feature vector for the entire image or provided crop.
This might also include extracted features regarding other parts of the image that are not relevant, while OpenSeg outputs pixel-wise embeddings. 
We provide an ablation for the advantages of using OpenSeg in \cref{tab:mrec}.
Thus, limiting the collected features to the ones related to the object improves our results.
Consequently, from the selected top-k images we use OpenSeg to compute pixel-wise language-aligned features for object $i$
and we compute a global language-aligned embedding for the object by aggregating pixel-embeddings of the projected pixels $\bP_{i\frame}$ using average-pooling.
This step can be observed in box 2a. in \cref{fig:frame_select}.
\todo[inline]{check figure reference for new figure}

\boldparagraph{Relationship feature computation}
\todo[inline]{bigger discussion on why we use BLIP?}
Similar to extracting per-object features, we also want to extract a global language-aligned feature embedding for relationships between two objects.
Again we make use of VLM and decide to use InstructBLIP~\cite{dai2023instructblip} since we identify in \cref{sec:related_compositional} that CLIP-like models are ill-suited to express compositional knowledge. 
Thus we use a BLIP-like model which visual feature embedding can be grounded with language to attend to the desired subjects.
Given the top-k images where both object instance $i$ and $j$ are visible, we crop the image to the union of their respective bounding boxes $box_{ij}^\frame = box_{i\frame} \cup box_{j\frame}$. 
Then, we encode the crop at multiple scales using the BLIP image encoder from InstructBLIP to align the features with the InstructBLIP language model. Providing multiple scales of the same crop has been shown beneficial to provide important context information in \cite{kerr2023lerf} and \cite{takmaz2023openmask3d}. The embedded crops are then aggregated using average-pooling. This step can be observed in \cref{fig:frame_select} box 
 2b.\looseness=-1 
 \todo[inline]{check figure reference for new figure}

\boldparagraph{Feature aggregation}
To provide a more robust and view-independent visual feature for objects and relationships, we average-pool all the global object features, and all the global relationship features previously extracted from each of the top-k frames. This results in two new global robust visual features: $\bff^\twoD_{o,i}$ for object $i$, and $\bff^\twoD_{r,ij}$ for relationship between objects $i$ and $j$. The set of all the object features and relationship features are denoted by
$\Fo{\twoD} = \{\bff^\twoD_{o,1}, ..., \bff^\twoD_{o,N}\}$ and $\Fr{\twoD}=\{\bff^\twoD_{r,1}, ..., \bff^\twoD_{r,M}\}$ respectively. 

\subsection{Graph distillation}
\label{sec:distillation}
The projected 2D object $\Fo{\twoD}$ and predicate $\Fr{\twoD}$ features can be directly used to predict 3D scene graphs if camera pose, depth and color images are available. 
However, in some circumstances, only 3D meshes or point clouds are provided. 
Furthermore, the fused 2D features can suffer from occlusions or prediction inconsistencies, resulting in noisy features. 
Therefore, we choose to distill the knowledge of the 2D vision-language models into a 3D network that operates on point clouds. 
To the best of our knowledge, the most suitable way to predict scene graph from 3D data is to leverage a GNN architecture.

Specifically, given a point cloud $\cP$, we construct a graph $\cG$ as defined in \cref{sec:graph_construction}. 
We use the GNN architecture with message passing as proposed in \cite{Wald_2020_CVPR} to output vision-language-aligned object node features as \mbox{$\Fo{\threeD} = \{\bff^\threeD_{o,1}, ..., \bff^\threeD_{o,N}\}$} with $N$ being the number of nodes, and relationship edge encoding features as \mbox{$\Fr{\threeD} = \{\bff^\threeD_{r,1},...,\bff^\threeD_{r,M}\}$} with $M$ being the number of edges.
 \todo[inline]{describe number of layers and projectors}

To enforce the vision-language alignment for our 3D graph features, we define a training objective using a cosine similarity loss between the 2D vision-language features and the 3D features for nodes and edges
\begin{equation}
    \cL = 1-\cos(\Fo{\twoD},\Fo{\threeD}) + 1-\cos(\Fr{\twoD},\Fr{\threeD}).
    \label{eq:distillation}
\end{equation}
Using this training objective, we distill the broad knowledge from the 2D vision-language foundation models into our 3D GNN. The process is depicted in \cref{fig:main_method}.

After the distillation, the 3D graph features live in the same embedding space of the 2D vision-language foundation models.

\subsection{Prediction and filtering}
\label{sec:prediction}

\boldparagraph{2D-3D Feature fusion} 
At inference time, we can perform open-vocabulary 3D scene graph prediction using only the distilled 3D features. 
However, if 2D images are available, we choose to fuse the 2D and 3D features in $\bff^{\twoD\threeD}_{o,i}$ and $\bff^{\twoD\threeD}_{r,ij}$ by average pooling the two for each feature pair \twoD-\threeD. This is inspired by Peng~\etal who observed in \cite{Peng2023OpenScene} that 2D features are beneficial to predict small objects, while 3D features yield good predictions for large objects with distinctive shapes. 
From this 2D-3D ensemble, we can infer node object classes and inter-object relationships in a two-step manner. 
First, we predict the object class of each node, and then using the inferred object classes we predict the relationship label on the edge between the classes.

\boldparagraph{Node prediction} 
As the first step to predict full open-vocabulary 3D scene graphs, we infer the object class label of each node from an open-vocabulary of arbitrary text prompts. 
These text prompts are encoded using the CLIP \cite{radford_2021_PMLR} text encoder to get the text features $\cT=\{\bt_1,...,\bt_N\}$, which are aligned with the OpenSeg \cite{ghiasi_2022_eccv} vision model and where $N$ is the number of candidate classes. 
To classify the object class, we compute the cosine similarity between the candidate text prompts and the 2D-3D ensemble graph embedding and choose the class with the highest similarity score to the node feature:
\begin{equation}
    \argmax_{n} \cos(\bff^{\twoD\threeD}_{o,i}, \bt_n).
    \label{eq:cosine_argmax}
\end{equation}

\boldparagraph{Relationship prediction}
Following the prediction of the node classes in an open-vocabulary manner, the second step predicts relationships informed by the object predictions from the first step.

Contrastive vision-language models such as CLIP \cite{radford_2021_PMLR} have been shown to have a poor compositional understanding of the world \cite{yuksekgonul2023when,yamada2022lemons,Ma_2023_CVPR,hsieh2023sugarcrepe} resulting in limited accuracy when used for tasks such as relationship prediction. 
Thus, querying predicates for the scene graph edges in a similar manner as we have done for our node prediction will yield poor results. 
We provide experimental results to this hypothesis in the \cref{tab:main_evaluation}.

To solve this issue we exploit generative VLMs, which are grounded via a specific task. 
These models usually produce outputs that perform better on VQA benchmarks or benchmarks where it is required to have compositional reasoning \cite{dai2023instructblip,li2023blip,pmlr-v162-li22n}. 
\todo[inline]{provide hypothesis why these models perform better}
However, a big drawback of deploying a generative approach is that restricting the output to a desired answer is not straightforward.
To this end, InstructBLIP~\cite{dai2023instructblip} is uniquely designed to give more output control using prompting. 
The InstructBLIP model consists of a vision transformer (ViT) encoder followed by a \textit{Qformer}, which receives context from learnable tokens, a user prompt and the output of the ViT. 
The \textit{Qformer} fuses and projects this information to the token space of a pre-trained LLM which is again conditioned on a user prompt.

We change the input to the \textit{Qformer} such that instead of receiving the vision features from the ViT, we provide our 2D-3D distilled ensemble features from \cref{sec:distillation} coming from our graph neural network.
To infer an accurate relationship grounded for a specific subject-object relation, we use the object class predictions from the first step to refine a template query to only output a relationship description for these two objects. 

The output of the scene-conditioned \textit{Qformer} is fed into the LLM which
is prompted to output a relationship description for the subject-object pair in the graph, given the same conditioned query.
This process is done in parallel for all edges in the scene graph to predict relationships for all subject-object pairs. 
The final result is an open-vocabulary 3D scene graph with open-vocabulary objects as well as open-vocabulary relationships.

\todo[inline]{open-vocab prompting via BERT?}

\section{Experiments}

\subsection{Experimental Setup}
\boldparagraph{Datasets} The choice of training data is generally fixed for other 3D scene graph methods. The 3DSSG dataset \cite{Wald_2020_CVPR} is at the time of writing this paper, the only dataset that provides semantic scene graph labels aligned with a 3D scene. This forces other methods \cite{Wald_2020_CVPR,Wu_2021_CVPR,koch2023lang3dsg,Zhang_2021_CVPR} to train and test on this rather small 3D dataset.
In contrast, our method can be trained independently from scene graph labels on a 3D dataset that provides a 3D representation with posed 2D images, including their depth. 
While 3DSSG provides high-quality 3D point clouds and scene graphs, the provided portrait images have a low FOV, leading to a suboptimal 2D feature extraction. 
Therefore, we choose ScanNet \cite{Dai_2017_CVPR}, a similar indoor dataset, which provides image frames with acceptable FOVs and high-quality point clouds. 
However, since 3DSSG is the only dataset to provide ground truth scene graph labels, we evaluate our distilled model quantitatively on it. \looseness=-1

\boldparagraph{Baseline methods}
Given the challenging nature of open-vocabulary 3D scene graph prediction, our method is the first true open-vocabulary 3D scene graph method, that not only models open-vocabulary objects, but also open-vocabulary relationships from 3D point clouds. 
Therefore, no comparable method exists. 
As the first open-vocabulary 3D scene graph prediction method we compare against the first closed-vocabulary semantic 3D scene graph estimation method 3DSSG \cite{Wald_2020_CVPR}. 
Further we compare against the current state-of-the-art~\cite{Wu_2021_CVPR,koch2023sgrec3d}.
Additionally, we devise some open-vocabulary baseline methods for a fair comparison of our method. 
The first baseline is a naive CLIP-based approach, where we try to predict relationships directly with CLIP \cite{radford_2021_PMLR}. The second baseline we propose is a CLIP-based alternative to our method, where we predict objects and predicates in a 2-step manner directly from 2D images, querying first objects and then relationships using CLIP. 
This baseline is meant to highlight the advantage of using InstructBLIP for relationship prediction. We also evaluate the performance of NegCLIP \cite{yuksekgonul2023when} which is supposed to have improved compositional understanding.
The third open-vocabulary baseline is similar to the concurrent work ConceptGraphs \cite{conceptgraphs} and utilizes a caption-based approach directly from 2D images. 
We use OpenSeg \cite{ghiasi_2022_eccv} and BLIPv2 \cite{li2023blip} to predict objects and their image captions, from which we extract objects and relationships for evaluation. 

For further insights into our devised baselines, the reader is referred to our supplementary work.
\todo[inline]{write about advantage of distillation}

\boldparagraph{Metrics}
Designing metrics to quantitatively evaluate the capabilities of open-vocabulary methods is a current problem. %
So far, the best approach remains evaluating an open-vocabulary method on closed-vocabulary metrics. 
In our case, we choose the commonly used top-k recall metric (R@k) \cite{Lu_2016_ECCV} for scene graphs. Following \cite{wald2022learning,Zhang_2021_CVPR,Wald_2020_CVPR,Xu_2017_CVPR,Yang_2018_ECCV}, we evaluate objects and predicates individually and relationships as subject-predicate-object triplets. 
Additionally, we provide a class-wise evaluation using the stricter mean recall metric (mR@k) \cite{Chen_2019_CVPR}. 
\todo[inline]{is it necessary to go into more detail here?}

\boldparagraph{Label mapping}
To evaluate our method on a fixed-vocabulary benchmark, we provide object text queries from the class label set of 3DSSG, which comprises 160 classes. 
We compute the cosine similarity and choose the top-k predictions based on their cosine similarities. 
However, since we predict relationships in a generative manner, we cannot provide fixed queries for our relationship prediction. 
The LLM will output the most likely and best descriptive relationship given the context as well as subject and object. 
To map this to the fixed label set, we employ BERT \cite{devlin2018bert}, a small language model with well-structured word embeddings. It encodes the output of the LLM and the target relationship labels set and computes the cosine similarity from which we select the \mbox{top-k} most likely candidates. 
We reason that BERT has a well-structured word embedding space and is a good look-up approach to finding the most fitting synonyms from the 27 relationship classes from the 3DSSG \cite{Wald_2020_CVPR} dataset, which contains spatial, supportive, semantic, and comparative relationships labels. 

\subsection{Closed-set 3D scene graph prediction}
\label{sec:closed_eval}
\vspace{-0.5em}
\boldparagraph{Comparisons with fully-supervised and zero-shot methods}
In \cref{tab:main_evaluation} we compare our new zero-shot open-vocabulary 3D scene graph prediction approach with both fully-supervised as well as other zero-shot baselines on the 3DSSG \cite{Wald_2020_CVPR} dataset. We outperform all our supervised baselines on object, predicate and relationship prediction. We demonstrate that a naive CLIP-based approach is ill-suited for relationship prediction, but also a two-step approach similar to our method by combining OpenSeg \cite{ghiasi_2022_eccv} and CLIP \cite{radford_2021_PMLR} or even NegCLIP \cite{yuksekgonul2023when} does not yield significant improvements. The caption-based approach also achieves considerably lower performances compared to our method. This is likely due to the poor quality of the 2D frames within the 3DSSG dataset, which negatively affects the caption-based approach which only uses 2D information for inference. In contrast, our approach uses a 2D-3D ensemble, where the distilled 3D features can compensate for the poor or missing 2D features.

Similar to other open-vocabulary approaches \cite{Peng2023OpenScene,takmaz2023openmask3d}, there is a noticeable gap to the state-of-the-art fully-supervised approaches. However, our zero-shot open-vocabulary approach is surprisingly competitive with the fully-supervised approach from a few years ago \cite{Wald_2020_CVPR}.

\figquality

\boldparagraph{Impact of class occurrence}
Fully-supervised methods are heavily biased by what they observe during training. 
Training samples of classes that are observed in a higher frequency are generally learned more effectively than rarer classes. 
In literature, there are multiple ways to alleviate this problem. 
Most scene graph methods \cite{Wald_2020_CVPR,koch2023sgrec3d,wang2023vl} for instance, uses a focal loss~\cite{Lin_2017_ICCV} to solve the problem of class imbalance in the training set. 
As a zero-shot approach, our method is less susceptible to class imbalance. 
To evaluate this, we compare in \cref{tab:headbodytail} the mR@k recall of our first open-vocabulary method with recent 3D scene graph methods
on the most common head classes, moderately common body classes and rare tail classes.
We observe that while fully supervised methods demonstrate impressive accuracy on common object and predicate classes, their recall drops drastically for rare tail classes. 
In contrast, our zero-shot method reports consistent results across all classes, achieving on-par results with current fully supervised methods for all object and predicate classes averaged and outperforming the fully supervised methods on tail-end object classes by a considerable margin. 
This demonstrates the core advantage of our zero-shot open-vocabulary approach that it performs robustly on a wide variety of objects and predicates.
\tabmaineval

\tabheadbodytailbaselines
\subsection{Ablation studies}
\noindent \textbf{Is our knowledge distillation effective?}
In the top part of \cref{tab:mrec} we ablate the effectiveness of the feature distillation from the VLMs to our graph neural network. 
We compare results on 3DSSG \cite{Wald_2020_CVPR} for our distilled 2D-3D ensemble method with a distilled 3D only method when posed images are not available and with a 2D only method where we directly use the 2D VLM features for 3D scene graph prediction. 
While the 2D method already shows good results, only when combining 2D and 3D features we reach the best performance of object and predicate prediction.

\noindent \textbf{What if we have ground truth objects?}
Our relationship prediction using the LLM from InstructBLIP is conditioned on the queried objects from the OpenSeg embedding. Therefore, the correctness of the relationship prediction is influenced by the accuracy of the object querying. To evaluate both modules decoupled from each other, we provide the ground truth labels to InstructBLIP from which the LLM predicts the relationship. In the bottom part of \cref{tab:mrec}, we observe that this has only a minimal impact, indicating that our method is robust towards slightly incorrectly predicted object nodes.

\noindent \textbf{What if we use CLIP instead of OpenSeg?}
We choose OpenSeg \cite{ghiasi_2022_eccv} as our 2D object feature extractor. A popular alternative is CLIP. In the bottom part of \cref{tab:mrec} we show experimentally that using OpenSeg as the 2D object feature extractor yields better results compared to CLIP.

\noindent \textbf{What if we learn predicates supervised?}
While the 3DSSG \cite{Wald_2020_CVPR} contains over 160 annotated object classes, the number of categorized predicates is below 50 and most related works only evaluate on 27 or fewer distinct predicates \cite{Wald_2020_CVPR,koch2023sgrec3d,Zhang_2021_CVPR,Wu_2021_CVPR}. Therefore, given the comparably small vocabulary of predicates, we choose to fine-tune our model on 27 fixed predicate classes with only a few labels per class (\mbox{\raisebox{-0.8ex}{\~{}}}100). In the bottom part of \cref{tab:mrec}, we observe that fine-tuning on 3DSSG improves predicate prediction with our model. Additionally, we observe synergy effects for object prediction. Hence, our VLM distillation training can also be an effective pre-training strategy when labels are scarce.

\vspace{-0.0cm}
\subsection{Qualitative Results}
In \cref{fig:qualitative_results}, we provide qualitative results from our open-vocabulary 3D scene graph prediction approach for two different scenes from ScanNet \cite{Dai_2017_CVPR}. We show the top-1 prediction for nodes and edges but filter edges where objects are further apart than 0.5m. The predicted object class labels are overall predicted correct and very specific, such as \textit{microwave} or \textit{dining chair}. The relationships between objects are generally correct as well with a diverse set of predicates such as \textit{next to, attached to, under, above}. The advantages of our open-vocabulary prediction are especially good to see for the predictions such as "tv \textit{mounted on} wall" or "microwave \textit{build into} kitchen cabinet".

\tabablation
\subsection{Limitations}
The experiments conducted in this paper demonstrate the 
\filbreak
\noindent 
potential and advantages of open-vocabulary 3D scene graph methods. We observe that while predicting open-vocabulary objects shows great potential, predicting open-vocabulary relationships remains a challenging problem.

Furthermore, the evaluation setup for systematically evaluating open-vocabulary 3D scene graph methods still remains an open problem. While closed-vocabulary evaluations are valuable, they cannot highlight the huge potential of open-vocabulary methods such as ours.
\section{Conclusion}
This paper introduces a new approach to learning semantic 3D scene graphs in an open-vocabulary manner from 3D point cloud data. Our method distills 2D VLMs into a 3D graph neural network thus creating a graph-based and language-aligned scene representation which can be queried and prompted to create an explicit open-vocabulary scene graph. To tackle the problem of lacking compositional knowledge in traditional VLMs, we split the relationship prediction into two steps, where we first query objects in a scene using CLIP and prompt relationships in a second step from the inferred objects using an LLM decoder.
Our proposed approach shows promising results when evaluated on a closed-set benchmark and qualitative results confirm the open-vocabulary nature of our method.
In future work, we see potential in improving relationship prediction even further to achieve even better and more reliable open-vocabulary 3D scene graph predictions that can be useful for many downstream tasks.

\newpage
\vspace{0.5em}\noindent\textbf{Acknowledgement}
 This work was partly supported by the EU Horizon 2020 research and innovation program under grant agreement No. 101017274 (DARKO).

{
    \small
    \bibliographystyle{ieeenat_fullname}
    \bibliography{main}
}
\setcounter{section}{0}
\setcounter{figure}{0}
\setcounter{table}{0}
\renewcommand\thesection{\Alph{section}}
\renewcommand\thetable{\Alph{table}}
\renewcommand\thefigure{\Alph{figure}}
\maketitlesupplementary

In this \textbf{supplementary material}, we first provide additional implementation details in \cref{supp:details}. 
Next, we detail our design choices for our open-vocabulary 3D scene graph approach in \cref{supp:design_choices}.
In \cref{supp:baselines} we provide additional details on our proposed baselines.
Next, we highlight the improved semantic understanding of our open-vocabulary method compared to fully-supervised methods in \cref{supp:semantics} 
and demonstrate the advantages with long-distance relationships compared to 2D-only open-vocabulary methods in \cref{supp:long_dist}. 
We show unique applications of how our open-vocabulary 3D scene graphs can be used in \cref{supp:application}.
Finally, we provide more qualitative results in \cref{supp:vis}.

\section{Implementation details}
\label{supp:details}
For our 3D graph backbone, we extract features from the point cloud using two PointNets that compute an initial 1024-dimensional feature vector for each node and edge. The graph features are refined using five layers of graph convolutions with message passing inspired by [44] and a hidden dimension of 2048. Finally, the node features are projected into the 768-dimensional CLIP space using a 5-layer MLP with ReLU activations and batch norm. The edge features are concatenated with the positional encoding from the BLIP-ViT and projected into the 1408-dimensional BLIP feature space using a 5-layer transformer architecture. The model is trained for 50 epochs using the Adam optimizer with weight decay, a learning rate of 5e-4, and a cyclic cosine-annealing learning rate scheduler. We use a batch size of 6 on a single Nvidia A100 GPU with mixed-precision. \looseness=-1

During inference time, we use the pre-trained CLIP ViT-L/14@336 text encoder to encode the object queries and a pre-trained Vicuna 7B LLM model from Hugging Face \footnote{\url{https://huggingface.co/Salesforce/instructblip-vicuna-7b}} for predicate prediction. 
To query the CLIP text encoder we use object classes from the 160 class label set from 3DSSG [44], but we are not limited to those and can also query other arbitrary object classes or even concepts rather than discrete classes. To prompt the LLM we design an open-ended prompt to get the most open-vocabulary response: ``Describe the relationship between [\textit{object1}] and [\textit{object2}]?". Here \textit{object1} and \textit{object2} are the object classes queried in the first step by CLIP. It is also possible to ask whether a specific relationship exists. However, we observe that providing more than five options confuses the LLM.
To map the LLM predictions to the closed-vocabulary benchmark label set, we use the bert-base-uncased model from Hugging Face \footnote{\url{https://huggingface.co/bert-base-uncased}} with 768-dimensional feature embeddings.

\section{Design choices}
\label{supp:design_choices}
To succeed with distilling an open-vocabulary 3D scene graph method from 2D foundation models, we first study which model and which dataset is best suited for the distillation.

\boldparagraph{Compositionality pilot-study}
Our approach highly depends on the knowledge encoded in the 2D vision-language model. However, Yuksekgonul \etal [52] and others [50] have demonstrated that current contrastive pre-trained vision-language models behave like bag-of-words models and have little understanding of compositionality. To evaluate whether a contrastively pre-trained VLM is suited for the distillation into our 3D scene graph model, we perform a pilot-study on a subset of the VL-Checklist Relation [52] benchmark. Differently from the evaluations conducted in [52], we do not evaluate whether the VLM can differentiate between the correct and incorrect relationship description but provide a set of queries where the VLM has to choose the most likely. This makes the task much harder for the VLM as the likelihood that the VLM picks the correct caption among the incorrect captions by random chance is much smaller. 
In the evaluation, we query the VLM using the query template ``A relationship of a [\textit{subject}] is [\textit{predicate}] a [\textit{object}]", where \textit{subject} and \textit{object} are fixed to the ground truth to solely evaluate the relationship understanding of the VLM.
We report the top-1, top-2, and top-5 recall scores denoting whether the correct predicate was in the top-k highest similarity scores.

\tabvlcheck

\noindent As expected, while CLIP [33], NegCLIP [52], and SigLIP [54] are exceptional zero-shot classifiers of objects, they cannot model inter-object relationships. The experimental evidence on a small controlled evaluation benchmark indicates that CLIP-like contrastively pre-trained VLMs do not have enough compositional knowledge about relationships that can be distilled into a 3D network. Therefore, in this paper, we choose to go beyond CLIP-like VLMs for relationship prediction and leverage a BLIP [7] vision encoder that can be projected into the token space of an LLM via a Qformer to predict relationships.

\begin{figure}[t]
    \centering
    \begin{subfigure}[c]{0.30\linewidth}
         \centering
         \includegraphics[angle=-90,origin=c,width=0.88\linewidth]{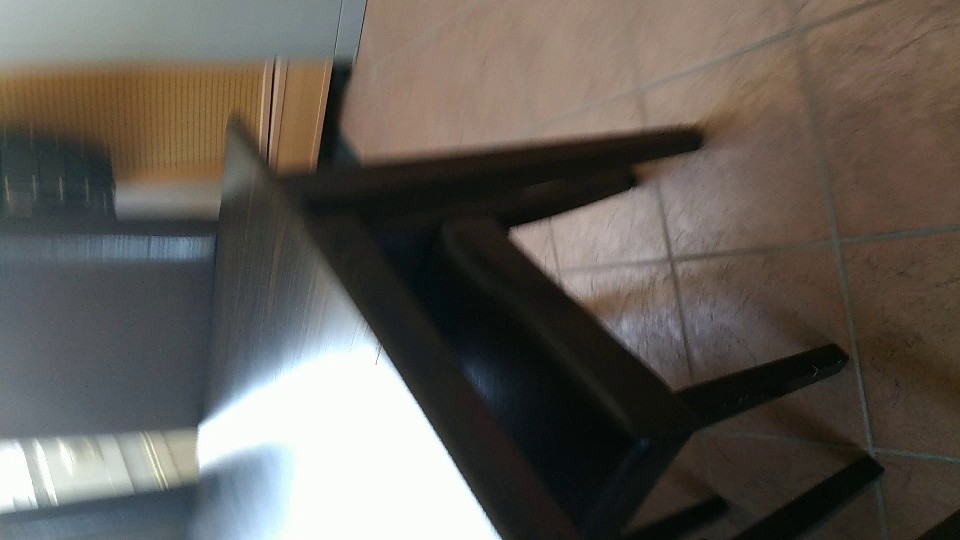}
     \end{subfigure}
     \begin{subfigure}[c]{0.30\linewidth}
         \centering
         \includegraphics[angle=-90,origin=c,width=0.88\linewidth]{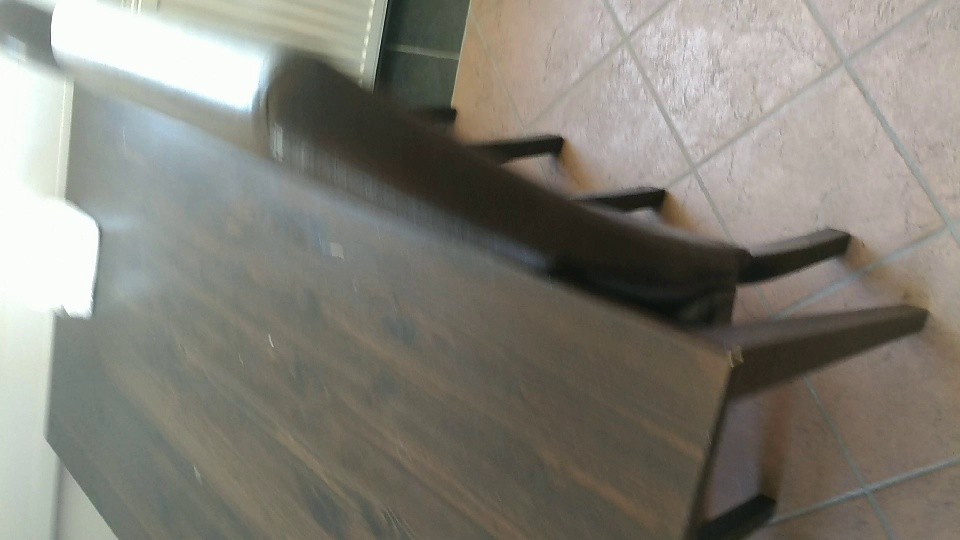}
     \end{subfigure}
     \begin{subfigure}[c]{0.30\linewidth}
         \centering
         \includegraphics[angle=-90,origin=c,width=0.88\linewidth]{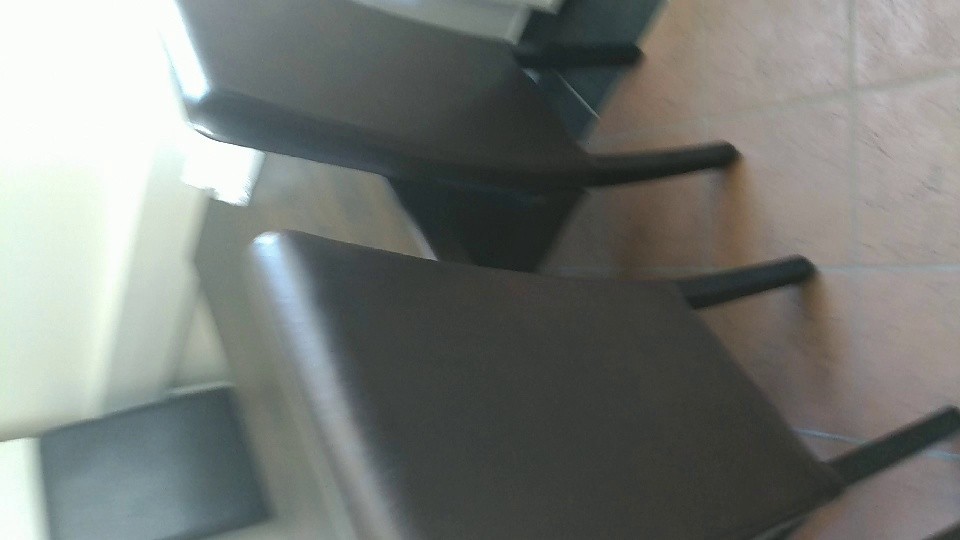}
     \end{subfigure}

     \ \\
     3RScan / 3DSSG\\
     \vspace*{0.5cm}

     \begin{subfigure}[c]{0.45\linewidth}
         \centering
         \includegraphics[angle=-0,origin=c,width=0.9\linewidth]{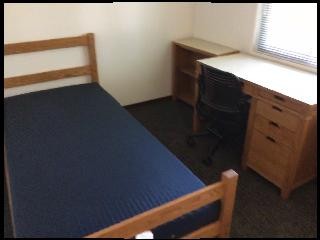}
     \end{subfigure}
     \begin{subfigure}[c]{0.45\linewidth}
         \centering
        \includegraphics[angle=-0,origin=c,width=0.9\linewidth]{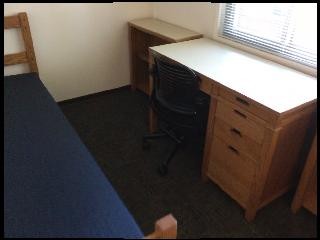}
     \end{subfigure}
     \begin{subfigure}[c]{0.45\linewidth}
         \centering
         \includegraphics[angle=-0,origin=c,width=0.9\linewidth]{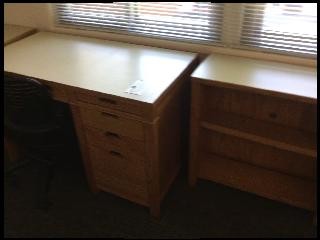}
     \end{subfigure}
     \begin{subfigure}[c]{0.45\linewidth}
         \centering
         \includegraphics[angle=-0,origin=c,width=0.9\linewidth]{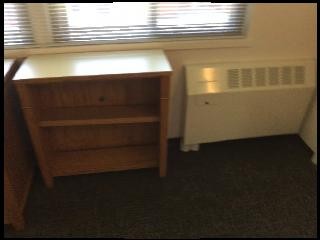}
     \end{subfigure}
    \ \\
     ScanNet
    \caption{\textbf{ScanNet vs. 3RScan.} We choose ScanNet over 3RScan / 3DSSG as a distillation dataset since the FOV of each frame is generally higher and more objects are visible in one frame.}
    \label{fig:scannet_vs_3rscan}
\end{figure}

\boldparagraph{Distillation dataset}
We choose to distill features on ScanNet [6] rather than 3RScan / 3DSSG [44], which we evaluate on. The reason for this is highlighted in \cref{fig:scannet_vs_3rscan}. Both datasets are indoor datasets depicting similar scenes. While ScanNet was recorded with an iPad with an attached depth sensor in landscape mode, 3RScan / 3DSSG was recorded with a Google Tango in portrait mode. The different recording setups result in entirely different vertical and horizontal field-of-views. We reason that to extract meaningful visual features representing the relationships between two objects, it is necessary that two objects are nearly fully visible in the same frame. This is rarely the case in 3RScan with its portrait setup. Therefore, we choose to use ScanNet for distillation as more of its frames depict more than one object.

\section{Baselines}
\label{supp:baselines}
In addition to proposing a novel open-vocabulary 3D scene graph prediction method, we also propose several baselines. Here we provide further details on these baselines.

\boldparagraph{CLIP (naive)} The most naive approach is to predict objects and predicates independently from each other directly using CLIP~[33].
We select images for each object instance as well as images where a pair of objects is shown similar to the process in Sec. 3.2 and encode them using the CLIP image encoder. Then we build a fully-connected graph from the encoded features and query the nodes with object class labels and the edges with predicate class labels.

\begin{figure*}[t]
    \centering
    \includegraphics[width=\textwidth]{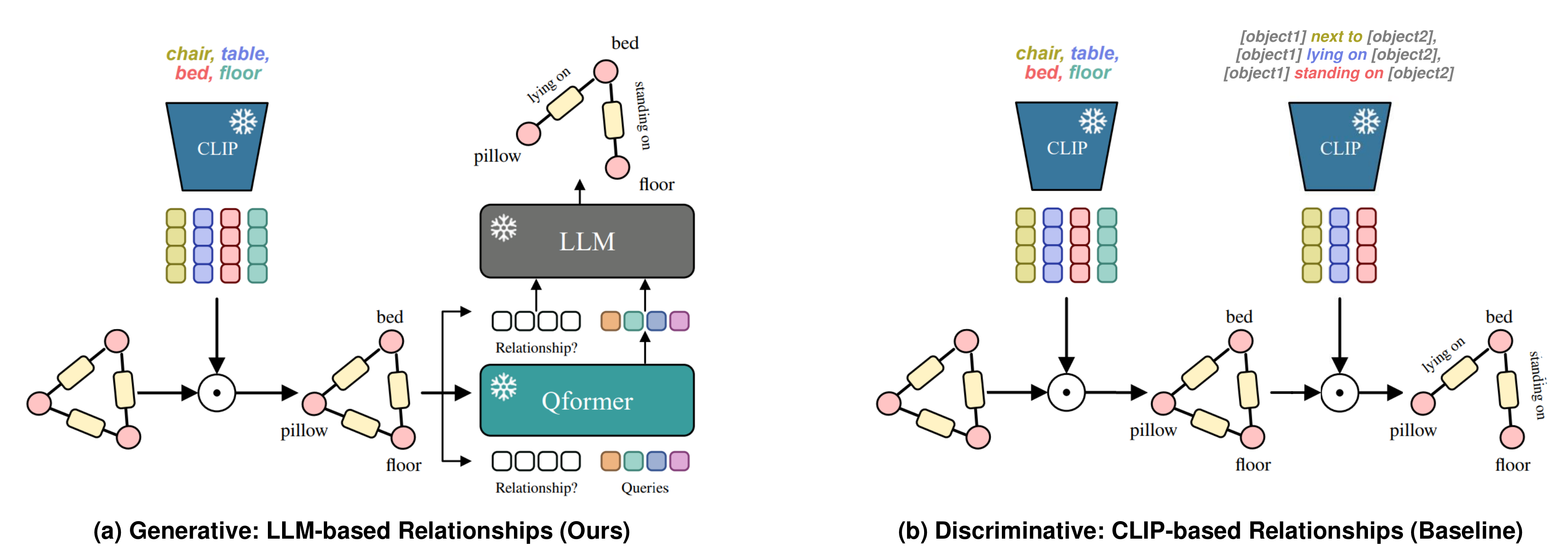}
    \caption{\textbf{Relationship prediction comparison.} We compare our generative relationship prediction approach using a prompted LLM (a), with a CLIP-based querying baseline (b) from Tab. 1. Due to the limited compositional knowledge of CLIP-like models, a discriminative approach where predicates can be directly queried performs much worse than a generative LLM-based approach.}
    \label{fig:clip_baseline}
\end{figure*}

\boldparagraph{CLIP \& NegCLIP} A more sophisticated approach using CLIP [33] or NegCLIP [52] is more similar to our two-step approach. The difference is shown in \cref{fig:clip_baseline}. Here we also first build a fully-connected feature graph and predict object classes by querying the class of each node. Then we use the predicted objects as context to query full relationships in a second step using CLIP. Using the predicted objects as context improves results compared to the naive approach, nevertheless, the results fall short of our LLM approach due to the limited compositional knowledge of both CLIP and NegCLIP.

\section{Improved semantics}
\label{supp:semantics}
While Tab. 1 in the main paper shows that our proposed open-vocabulary 3D scene graph method achieves overall worse performance compared to the current SOTA fully-supervised methods, Tab. 2 demonstrates the advantages of an open-vocabulary method, where we outperform the fully-supervised baselines on long-tail distribution classes. To give further insights into the benefits of our proposed open-vocabulary method, we provide scores on selected object and predicate classes in \cref{tab:class_study}.
\tabclasscompare
It shows that our open-vocabulary method outperforms the fully-supervised methods on very specific and semantically descriptive classes. For instance, for objects our network is better at differentiating a \textit{chair} from a \textit{dining chair} or a \textit{table} from a \textit{bedside table}. At the same time, fully-supervised methods, likely due to class imbalance during training, often only predict a generic class rather than the most specific class possible.
This is similar for predicates. While the fully-supervised methods generally perform well on all predicates, highly semantic and specific predicates such as \textit{covering} or \textit{belonging to} are predicted less accurately. In contrast, our open-vocabulary method performs particularly well on semantic predicates such as \textit{standing on}, \textit{covering} or \textit{belonging to}.

\section{Long distance relationships}
\label{supp:long_dist}
In Tab. 3, we provide an ablation for 3D scene graph prediction solely with 2D vision-language models. Only using 2D data performs worse than our learned 2D-3D ensemble approach. 

\begin{figure}[h]
    \centering
    \begin{subfigure}[c]{0.99\linewidth}
         \centering
         \includegraphics[width=\linewidth]{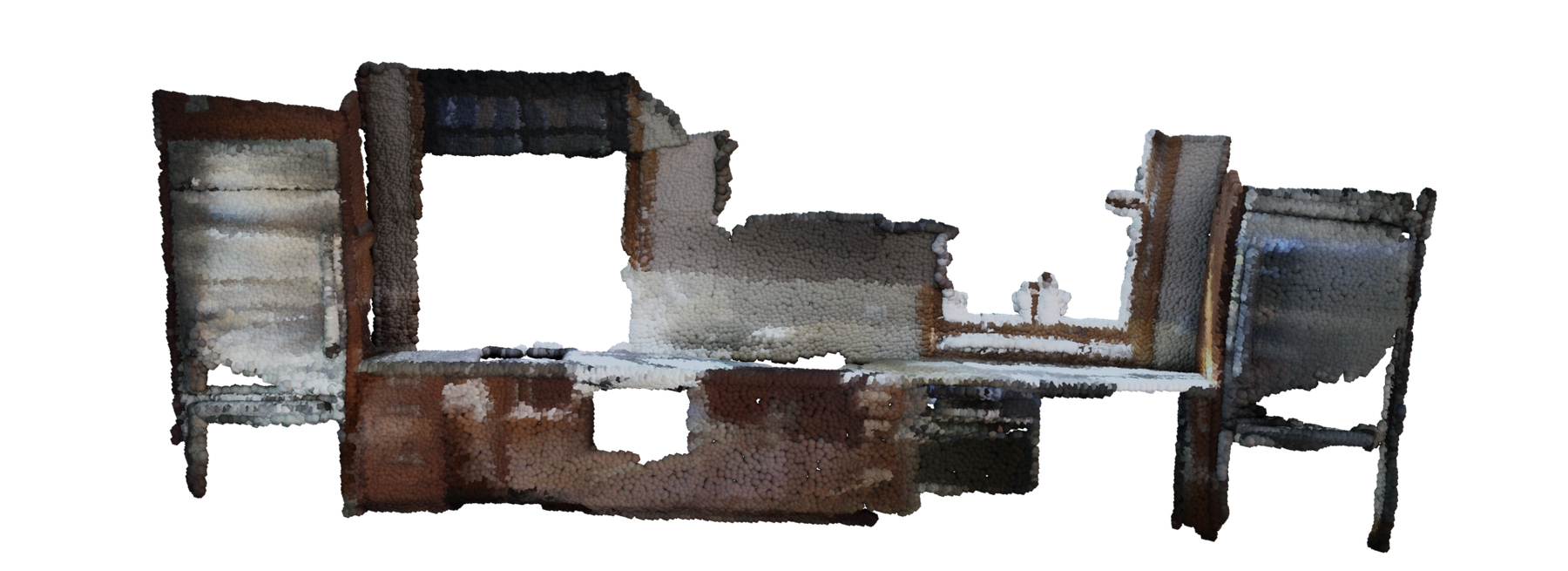}
     \end{subfigure}
     \begin{subfigure}[c]{0.99\linewidth}
         \centering
         \includegraphics[width=\linewidth]{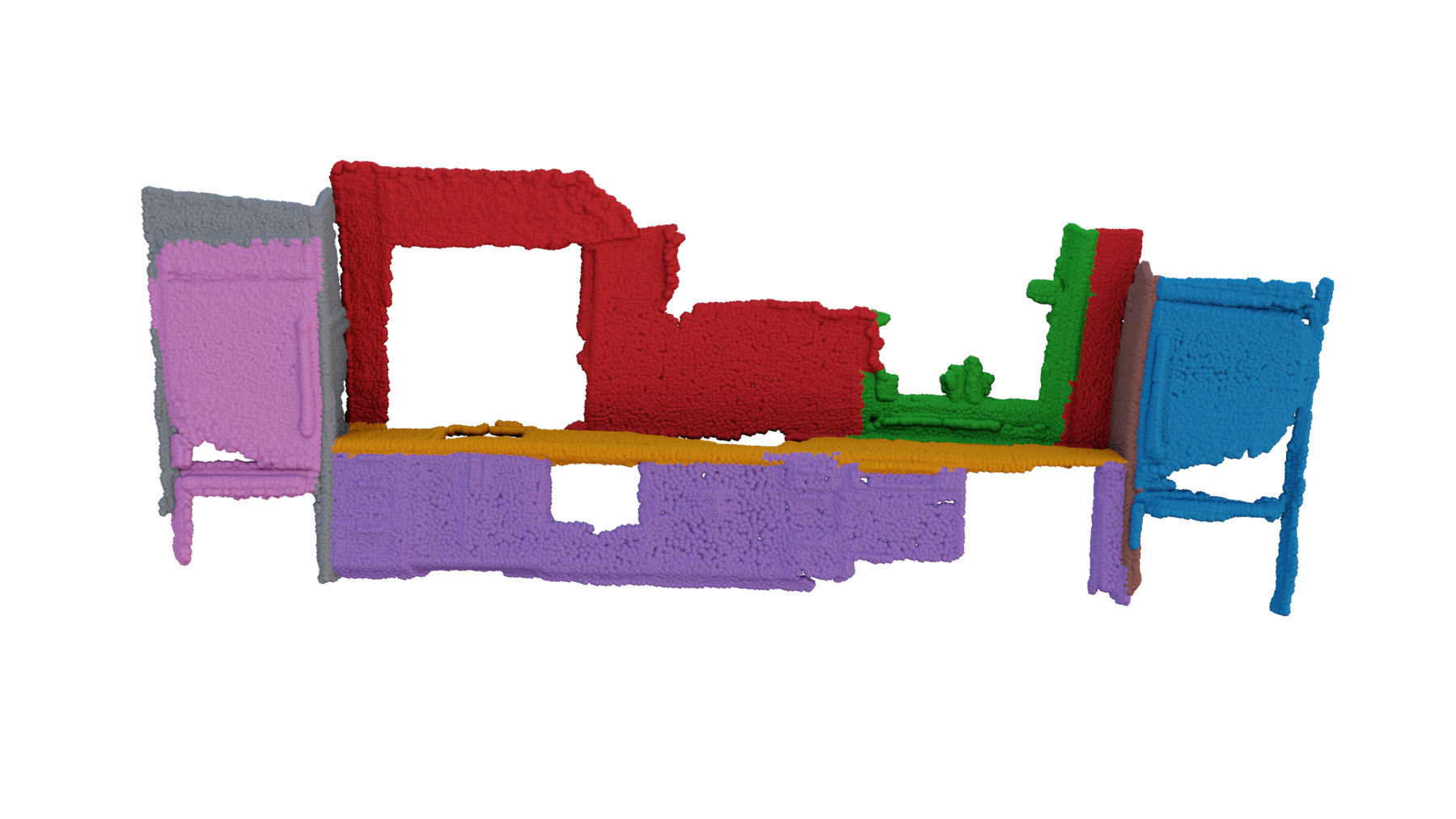}
     \end{subfigure}
     \begin{subfigure}[c]{0.80\linewidth}
         \centering
         \includegraphics[width=\linewidth]{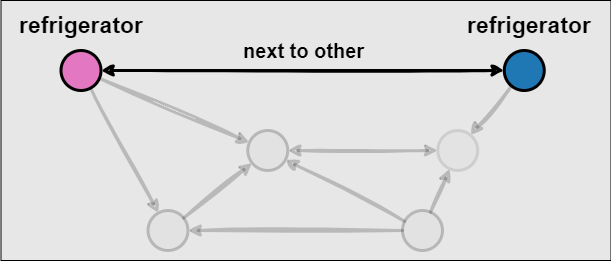}
     \end{subfigure}
    \caption{\textbf{Long distance relationships.} In contrast to a 2D-only relationship prediction approach, which requires two objects to be visible in an image together, our 3D approach can predict relationships for two arbitrary far objects.}
    \label{fig:elong_dist}
\end{figure}

While a prediction using 2D images is possible, a significant disadvantage of relying only on 2D data 
is that to predict a relationship between two objects, those two objects must be visible together in at least one frame. In contrast, our method does not have this limitation since it processes the 3D point cloud and can predict a relationship between two objects of arbitrary distance in a point cloud. \cref{fig:elong_dist} shows such two far-apart objects that are not close enough to appear in a shared frame, but still have a meaningful relationship detected by our method. 

\section{Applications}
\label{supp:application}
\subsection{3D Triplet localization}
3D scene graphs are useful for various downstream computer vision or robotics tasks. In \cref{fig:triplet_loc} we demonstrate one of those use cases uniquely suited to our language-aligned open-vocabulary 3D scene graphs. 
First, a 3D scene is encoded as an open-vocabulary 3D scene graph using our method. This representation is now queryable and promptable with an open vocabulary, making it a versatile tool for various scene understanding tasks. 
We demonstrate its usefulness for object localization in a 3D point cloud. Unlike other object localization methods [31], our goal is not to localize all objects of the same class but a specific instance that fits a relationship description. We encode a relationship description using the CLIP [33] and BERT [8] language encoders to generate a triplet feature representing the relationship. Then, we perform a subgraph-matching based on the cosine similarity of each triplet in the encoded scene graph with our target triplet feature. We select the triplet with the highest similarity score and reference it in the point cloud using the scene graph-point cloud alignment.

\figtripletloc

\subsection{Material prediction}
We present another application of zero-shot object attribute/material prediction, evaluated quantitatively in \cref{fig:material}. The material prediction can be performed without further training with the same querying strategy described in Sec. 3.4. Predicting attributes for each object further enriches the predicted 3D scene graph. We provide a top-1 accuracy metric comparison with OpenScene [31], a point cloud-based open-vocabulary method, on 3DSSG. 
\ours outperforms OpenScene for most materials and also achieves a higher average accuracy for all classes. Note however that OpenScene predicts the material per point while we predict the material per instance.
\figmaterial

\subsection{Reasoning over object affordances}
A further application is the reasoning over scene-specific affordances using \ours. Given the open-vocabulary representation computed by our method, we can prompt the LLM to predict affordances between objects. These affordances are grounded by the processed scene. In \cref{fig:weight_reasoning}, we demonstrate how \ours can reason over which objects can be picked up by a human by prompting the LLM "Can you lift [x] from [y]". Our model correctly predicts that the pillows can be picked up from the bed while the bed would be too heavily to lift from the carpet.
\figweightreasoning

\section{Additional 3D scene graph predictions}
\label{supp:vis}
\begin{figure*}[t]
    \centering
    \begin{subfigure}[c]{0.329\textwidth}
         \centering
         \includegraphics[width=\textwidth]{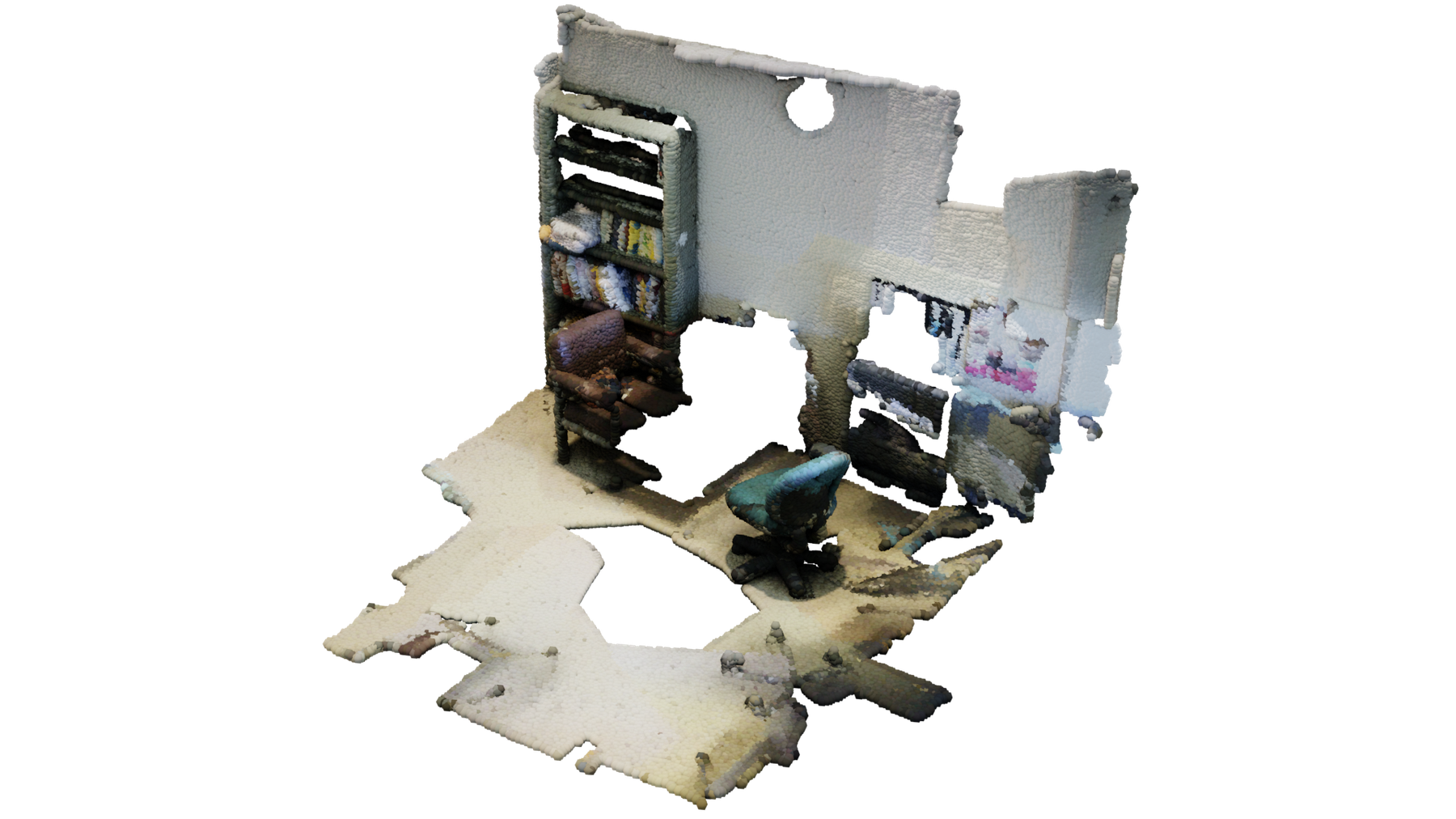}
     \end{subfigure}
     \begin{subfigure}[c]{0.329\textwidth}
         \centering
         \includegraphics[width=\textwidth]{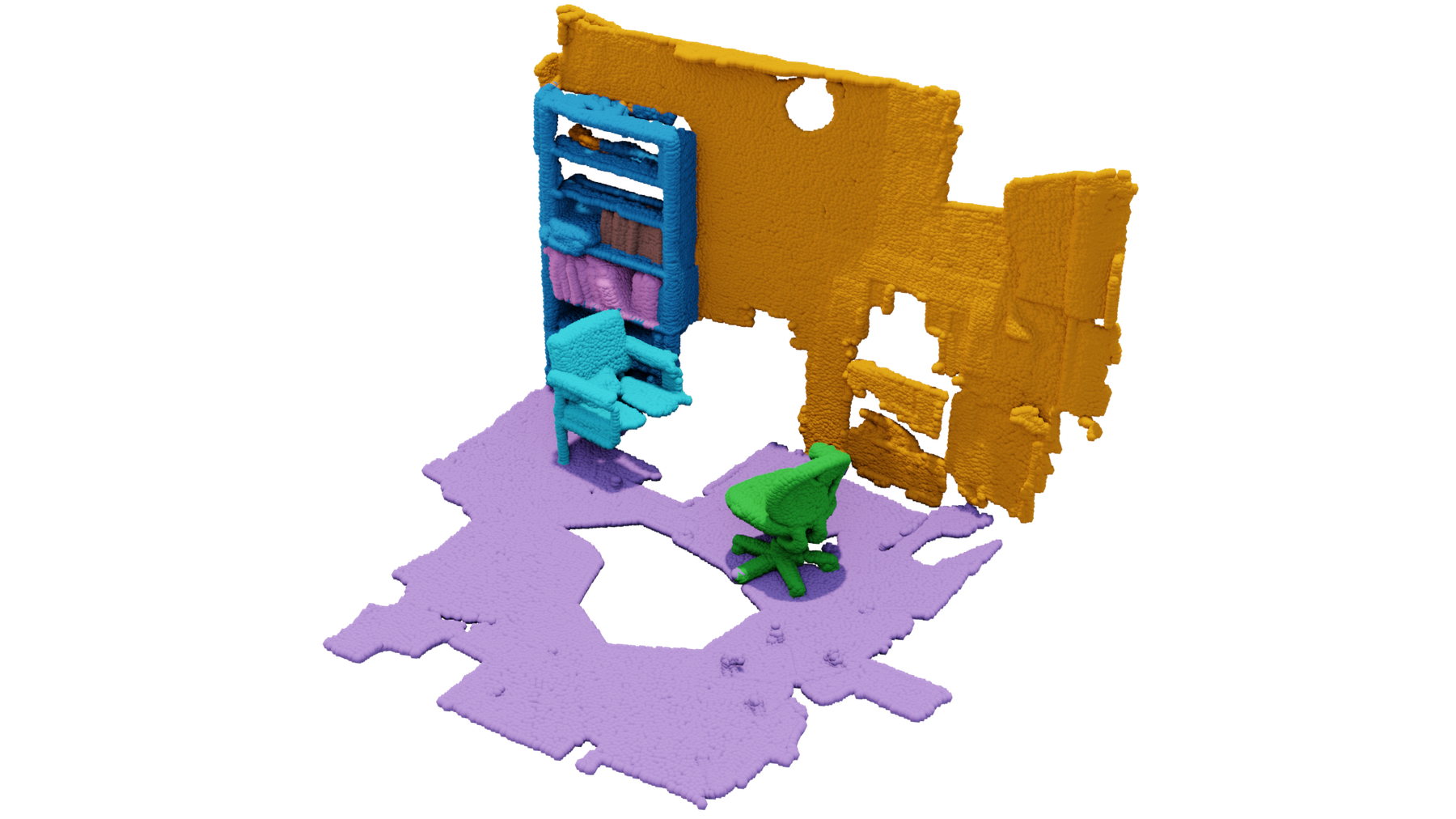}
     \end{subfigure}
     \begin{subfigure}[c]{0.329\textwidth}
         \centering
         \includegraphics[width=\textwidth]{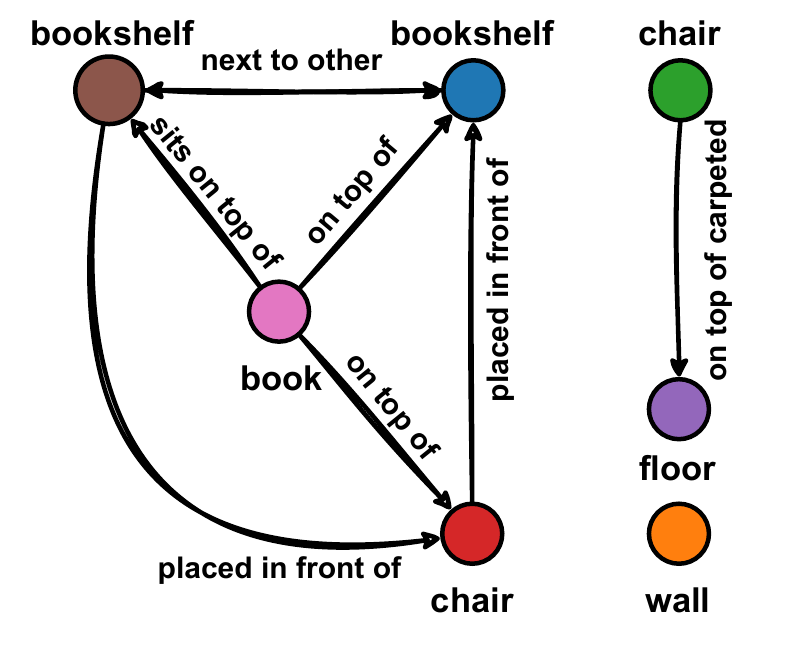}
     \end{subfigure}

     \begin{subfigure}[c]{0.329\textwidth}
         \centering
         \includegraphics[width=\textwidth]{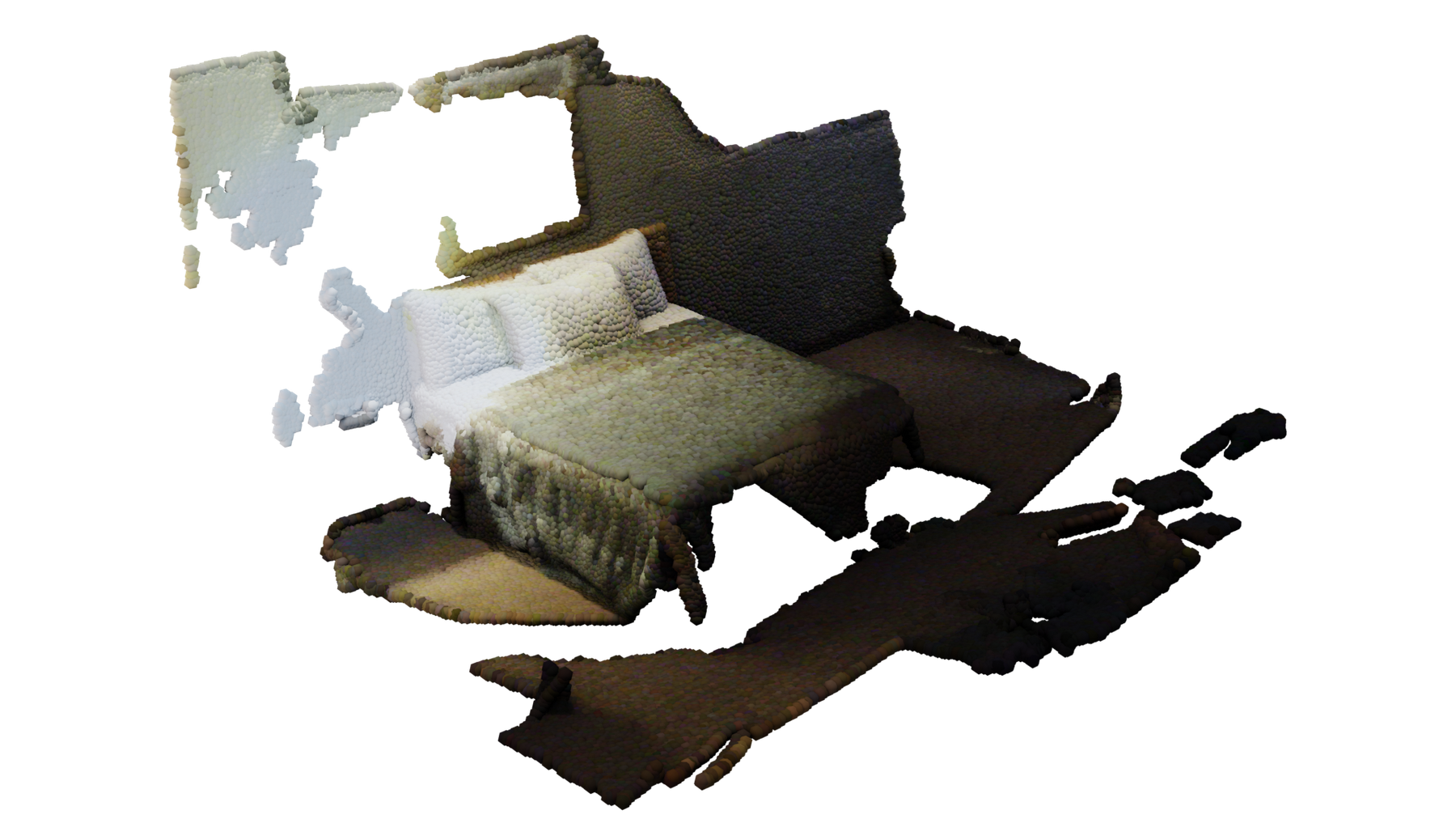}
     \end{subfigure}
     \begin{subfigure}[c]{0.329\textwidth}
         \centering
         \includegraphics[width=\textwidth]{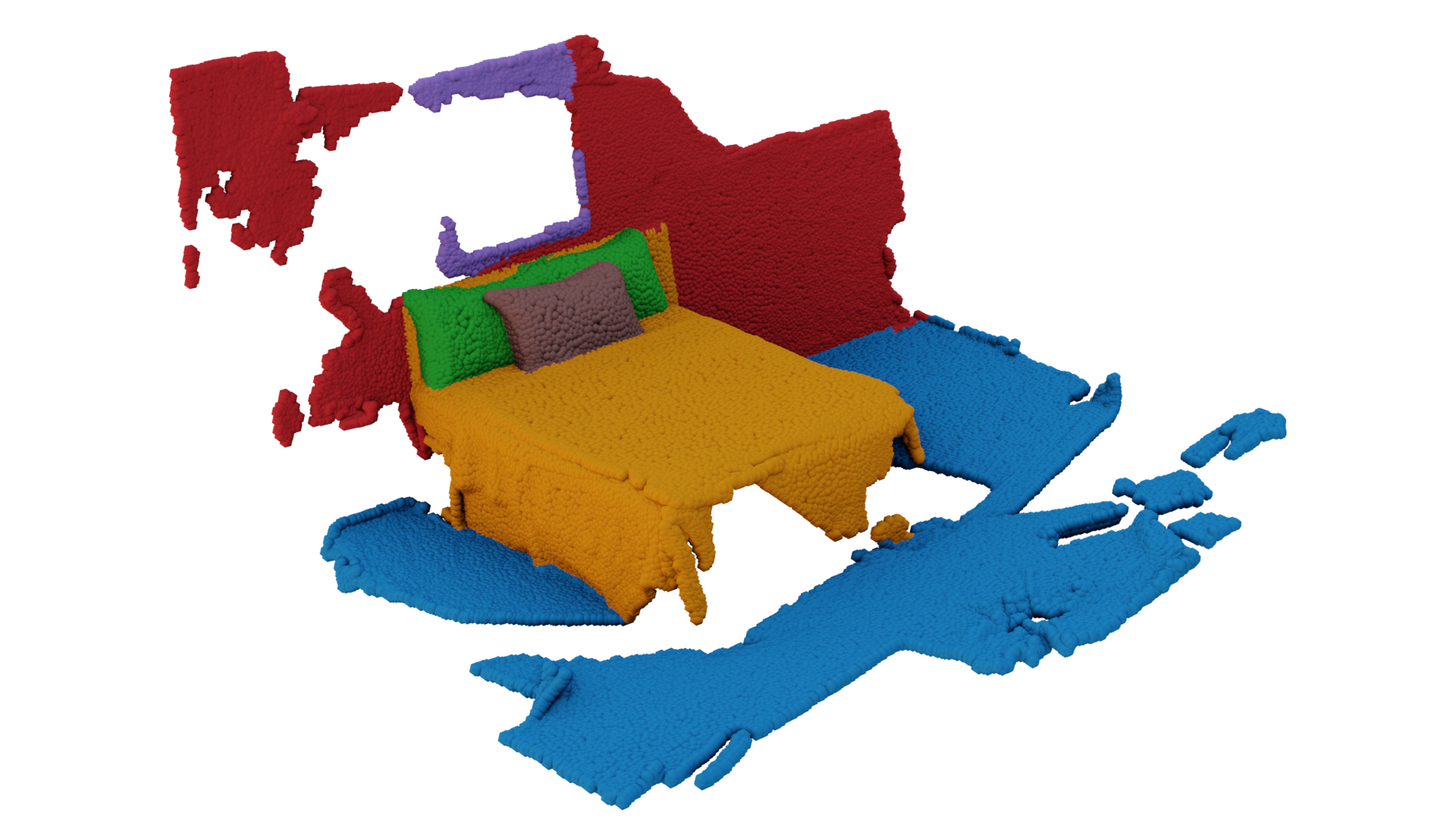}
     \end{subfigure}
     \begin{subfigure}[c]{0.329\textwidth}
         \centering
         \includegraphics[width=\textwidth]{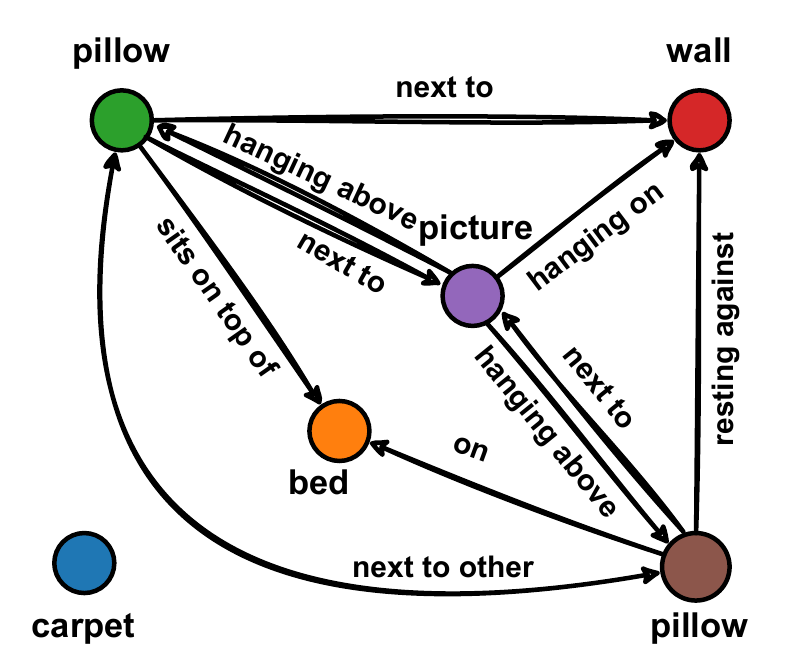}
     \end{subfigure}

    \begin{subfigure}[c]{0.329\textwidth}
         \centering
         \includegraphics[width=\textwidth]{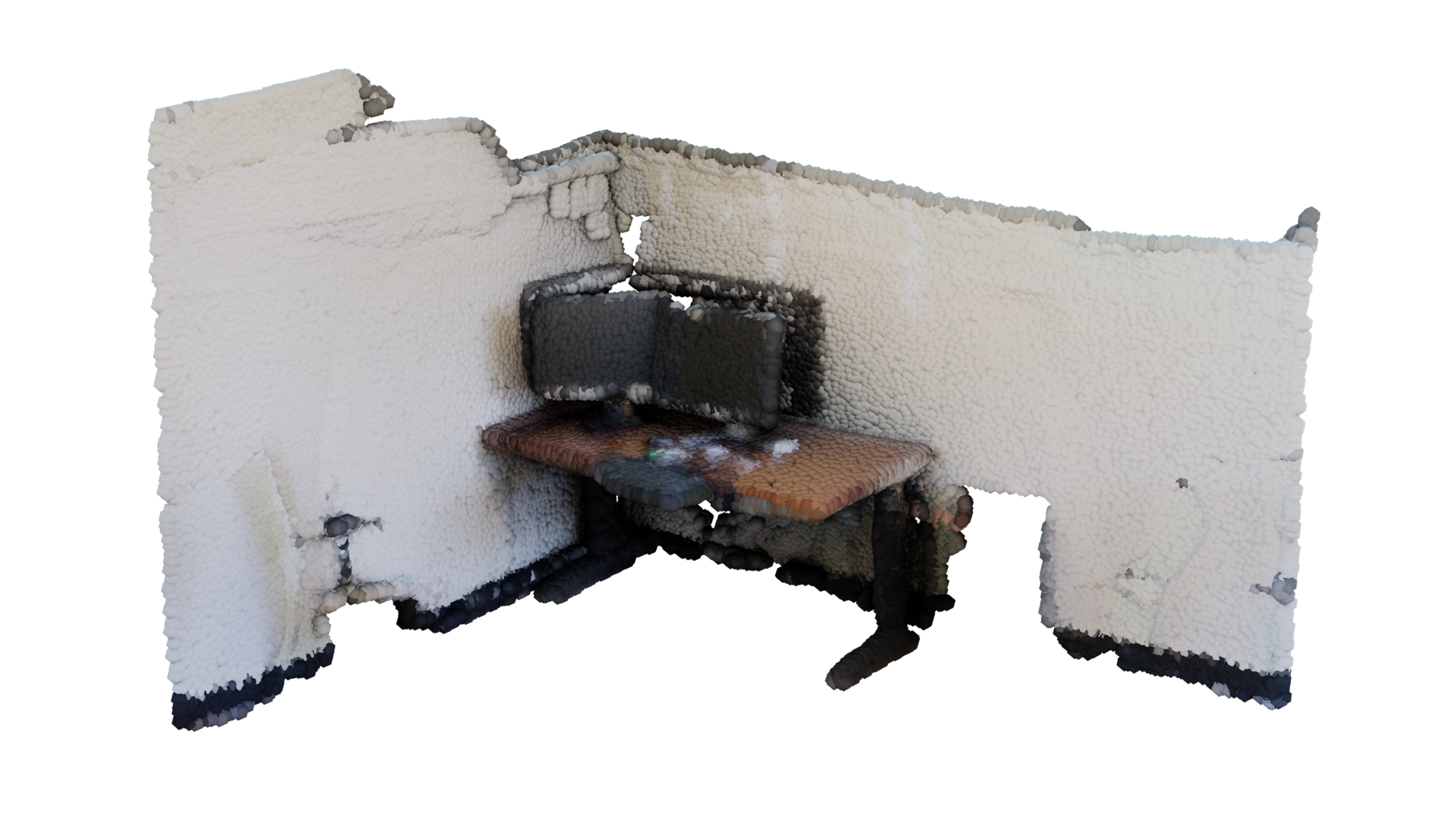}
     \end{subfigure}
     \begin{subfigure}[c]{0.329\textwidth}
         \centering
         \includegraphics[width=\textwidth]{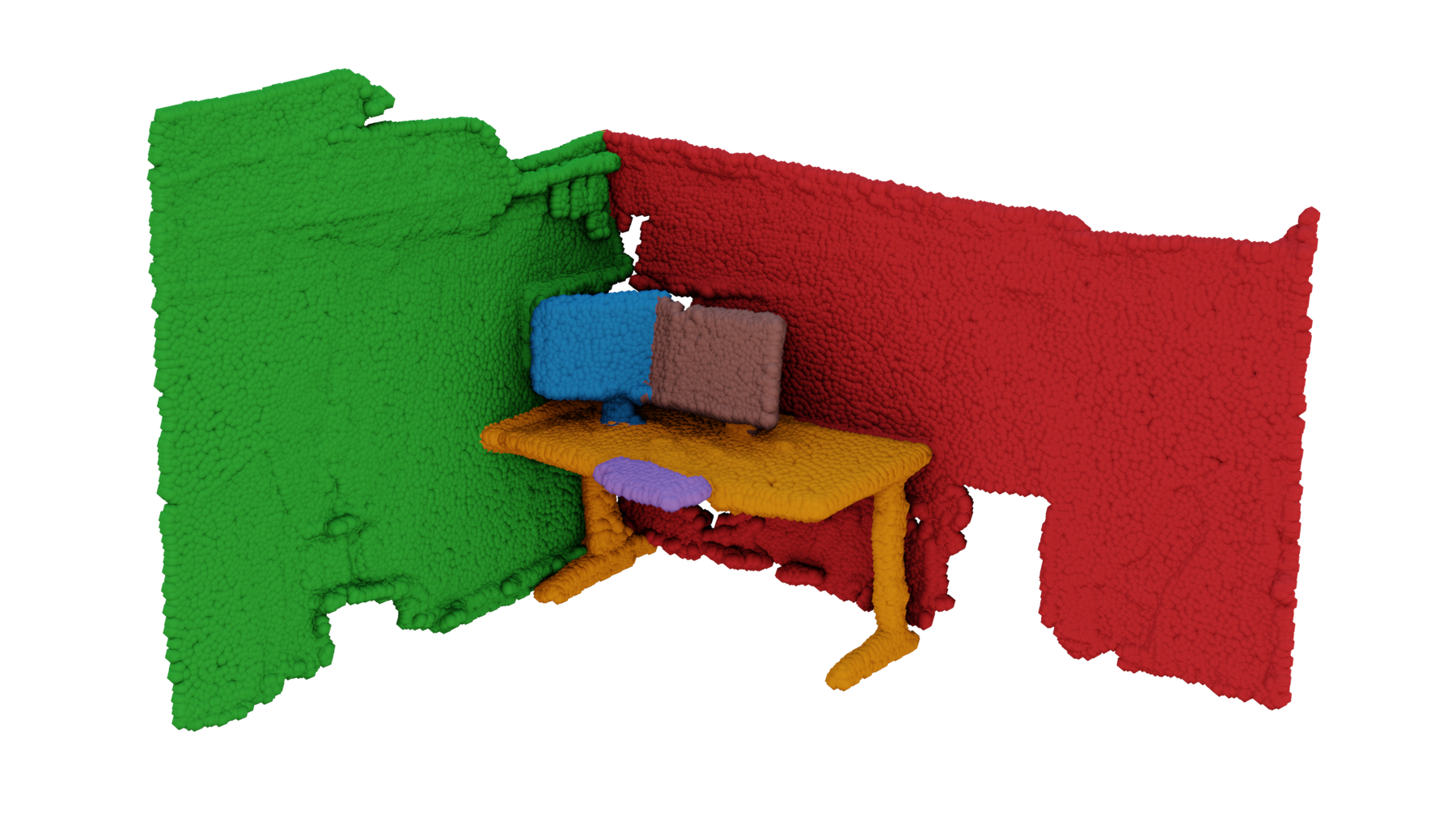}
     \end{subfigure}
     \begin{subfigure}[c]{0.329\textwidth}
         \centering
         \includegraphics[width=\textwidth]{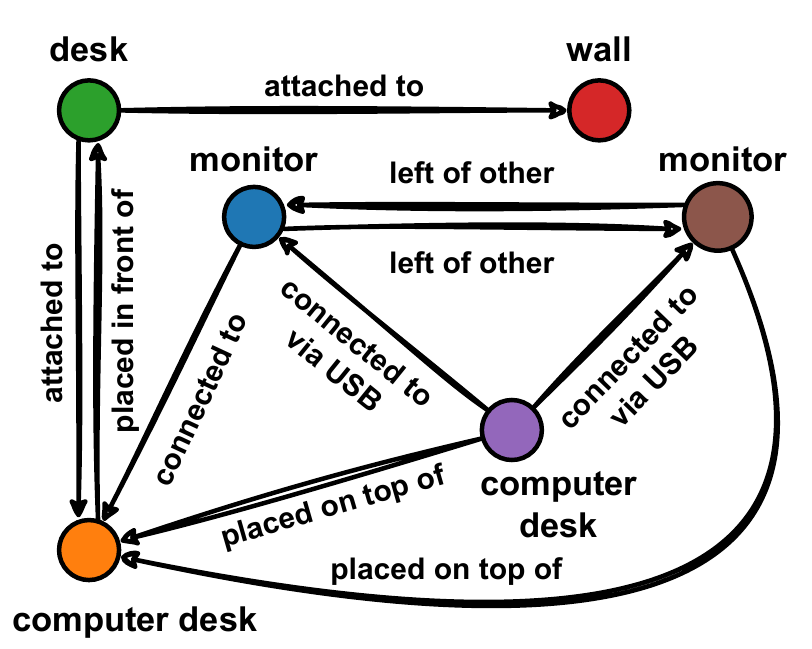}
     \end{subfigure}

     \begin{subfigure}[c]{0.329\textwidth}
         \centering
         \includegraphics[width=\textwidth]{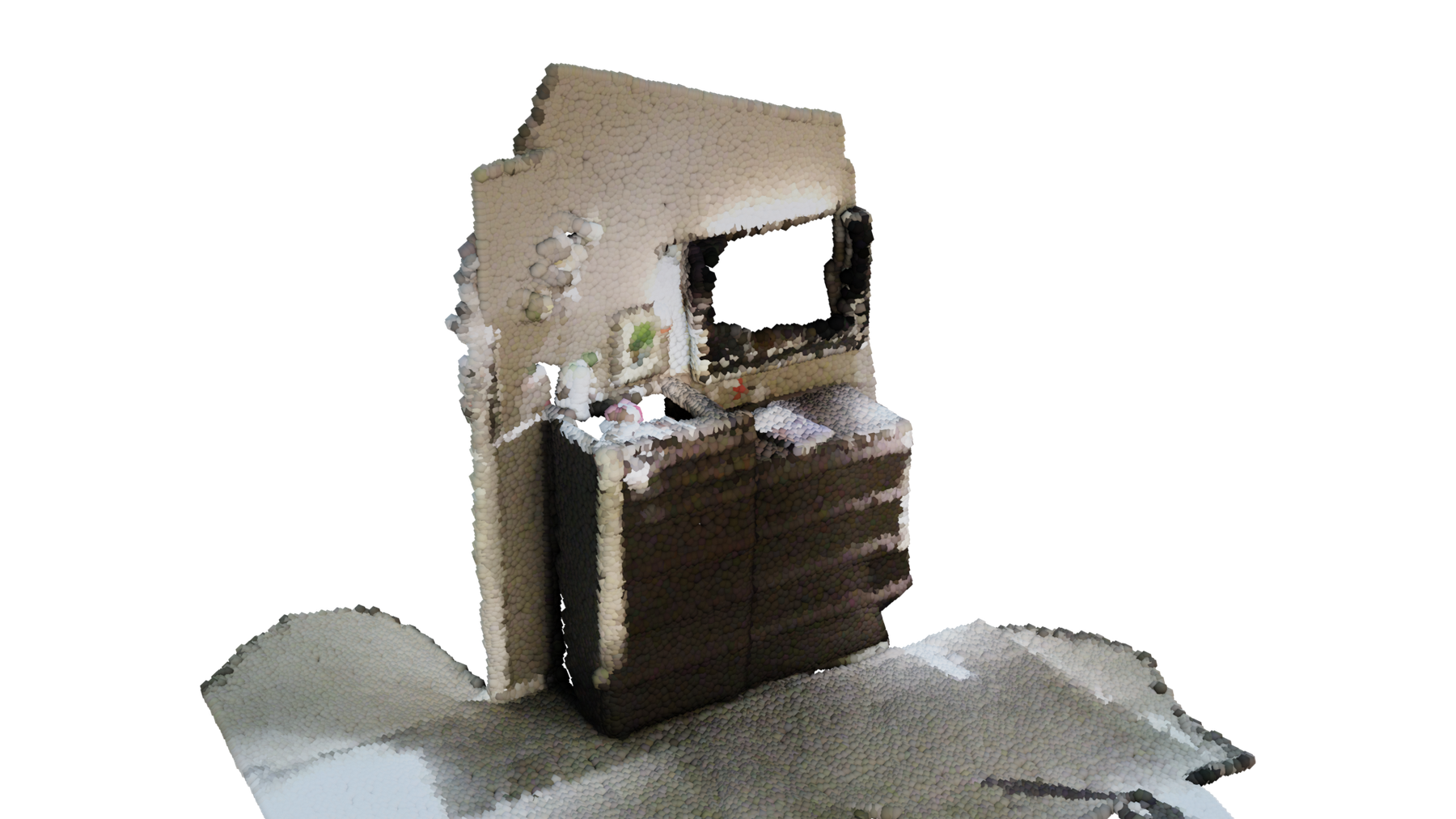}
     \end{subfigure}
     \begin{subfigure}[c]{0.329\textwidth}
         \centering
         \includegraphics[width=\textwidth]{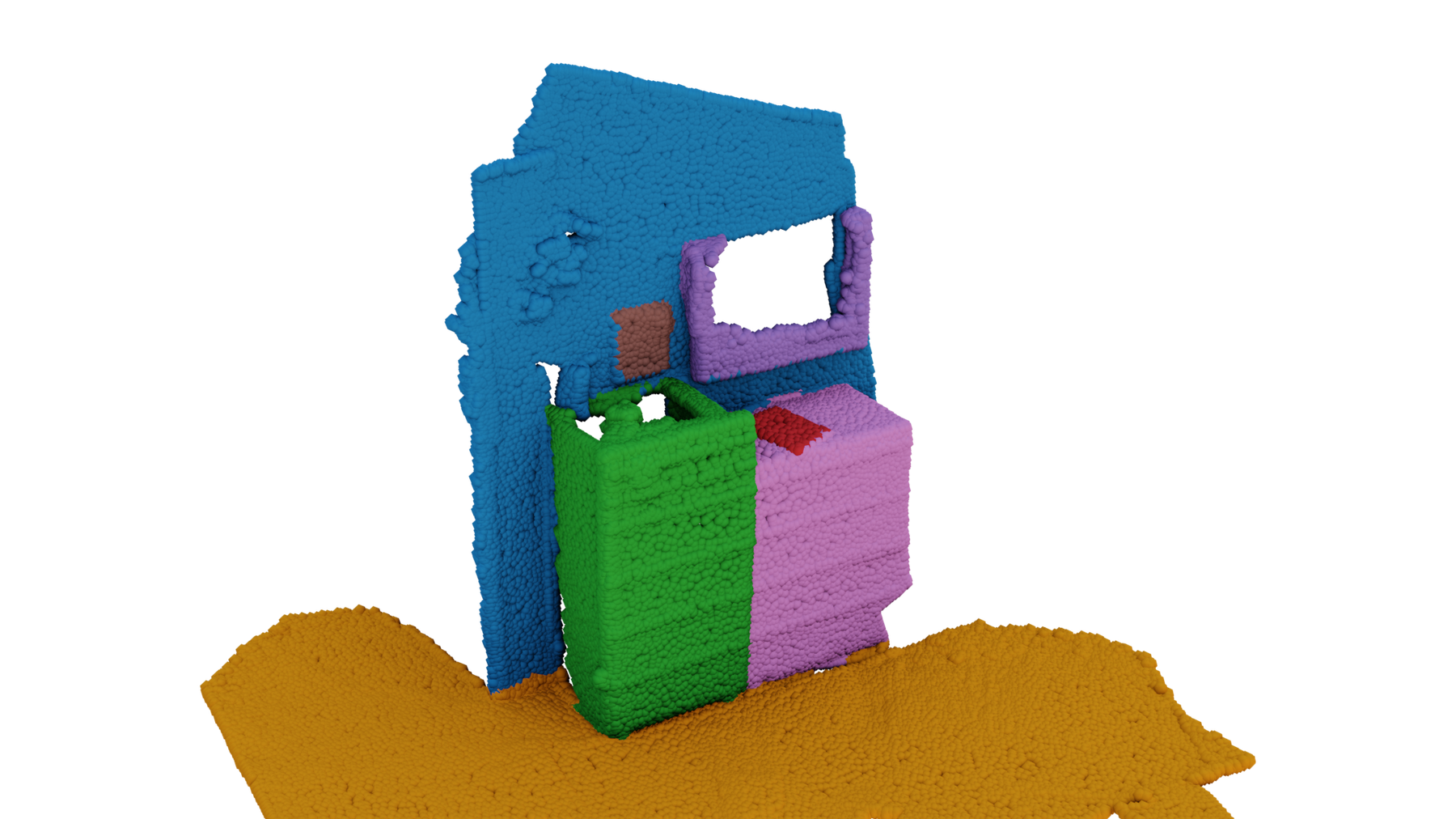}
     \end{subfigure}
     \begin{subfigure}[c]{0.329\textwidth}
         \centering
         \includegraphics[width=\textwidth]{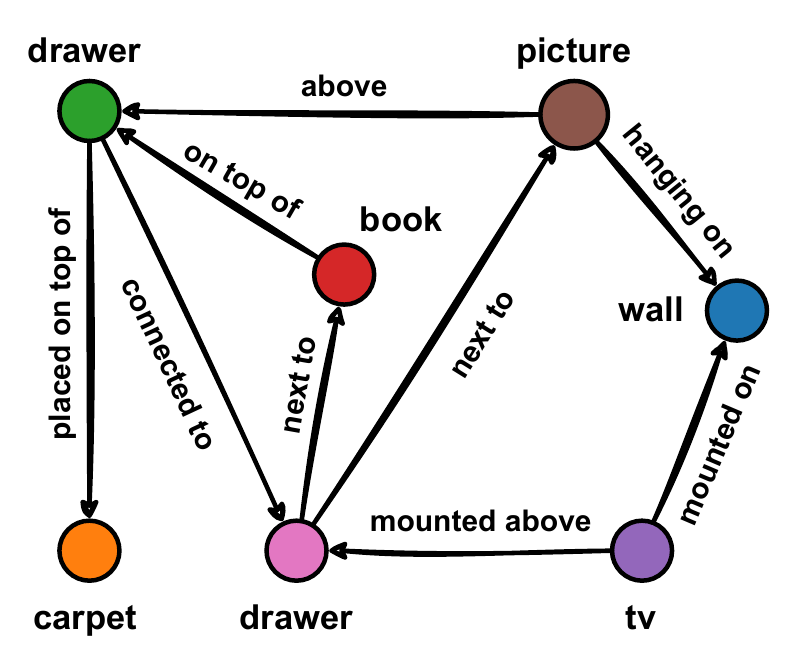}
     \end{subfigure}

    \caption{\textbf{Qualitative open-vocabulary 3D scene graph predictions.} Left: Colored point cloud input; Middle: Class-agnostic mask; Right: Predicted open-vocabulary 3D scene graph.}
    \label{fig:qualitative_results_supp}
    
\end{figure*}

In \cref{fig:qualitative_results_supp}, we provide additional 3D scene graph predictions on ScanNet [6]. 
Relationships for objects that are further apart than 0.5m are pruned for clarity in the visualization.
Overall, the 3D scene graph predictions are correct and the advantages of an open-vocabulary method become especially apparent for rare and specific object classes such as \textit{computer desk} or precise relationship descriptions such as \textit{tv mounted on wall}.
But our open-vocabulary approach still has several limitations, such as overall low diversity in the predicted relationships. However, this limitation is not unique to our open-vocabulary method but also remains an issue with the current state of fully-supervised methods.

Nonetheless, our approach also has unique limitations, such as LLM-typical hallucinations like \textit{computer desk (keyboard) connected via USB to monitor} or imperfect geometric understanding where two monitors are both predicted to be \textit{to the left of} each other.

\end{document}